\documentclass{article}
\usepackage{iclr2027_conference,times}
\usepackage[T1]{fontenc}

\usepackage{url}
\usepackage{graphicx}
\usepackage{amsmath,amssymb}
\usepackage{booktabs,multirow,array}
\usepackage{xcolor,colortbl}
\definecolor{availfill}{HTML}{F3F3F3}
\definecolor{missingc}{HTML}{2B2B2B}
\definecolor{flowc}{HTML}{2A9D4F}\definecolor{flowfill}{HTML}{E3F1E7}
\definecolor{worldc}{HTML}{2F6DB3}\definecolor{worldfill}{HTML}{E1EBF7}
\definecolor{mintc}{HTML}{C0392B}\definecolor{mintfill}{HTML}{FBE9E6}
\definecolor{bestcell}{HTML}{F6D9D5}
\usepackage{pifont}
\usepackage{tikz}
\usetikzlibrary{arrows.meta,positioning,calc,fit,backgrounds}
\usepackage{placeins,float}
\usepackage{subcaption}
\usepackage{hyperref}
\hypersetup{hidelinks}
\graphicspath{{figures/}}

\usepackage{wrapfig}

\newcommand{\mint}{MINT}
\newcommand{\mail}{MAIL-Bench}

\title{Learning to Act under Visual Interruptions with Vision-Language-Action Models}

\author{Mingle Jiang$^{1}$, Rui Xu$^{1}$, Yunke Wang$^{1}$, Chang Xu$^{1}$\\
\textnormal{$^{1}$School of Computer Science, The University of Sydney}}

\iclrfinalcopy

\begin{document}

\maketitle
\lhead{Preprint.}

\begin{abstract}
Vision-language-action (VLA) models have demonstrated strong capabilities in robotic manipulation, but they are typically developed and evaluated with all camera streams available throughout task execution. When a camera stops delivering frames during task execution, the policy must continue acting without access to subsequent observations from the missing view. Despite its practical importance, how such interruptions affect closed-loop manipulation remains insufficiently understood. To investigate this problem, we introduce \textbf{MAIL-Bench}, a benchmark that evaluates visual interruptions with VLA models.
By interrupting different cameras at multiple stages of each policy's successful reference trajectory, MAIL-Bench measures how well policies retain their capabilities when visual inputs become unavailable. 
Building on this benchmark, we propose \textbf{MINT}, which first trains VLA
policies to remain functional under missing visual inputs. At inference time,
MINT selectively supplements missing observations using optical-flow
extrapolation or an action-conditioned world model, and withdraws predicted
views when they become unreliable. Experiments on $\pi_{0.5}$ and GR00T~N1.5 show that MINT significantly improves task success under camera loss over the original models. Experiments on AgiBot~G2 further demonstrate the real-robot deployment under camera loss. The benchmark is available at \url{https://minglejiang.github.io/Mail-Bench/}.
\end{abstract}

\section{Introduction}
\label{sec:intro}
Vision-language-action (VLA) models
\citep{brohan2023rt2,octo2024,kim2024openvla,black2024pi0,black2025pi05,nvidia2025groot,xu2026stellavla}
use visual observations and language instructions to perform robotic
manipulation. Standard evaluation benchmarks
\citep{liu2023libero,mees2022calvin,james2020rlbench,nasiriany2026robocasa365}
typically assume that all camera streams remain available throughout task
execution. In practice, however, one or more camera streams may become
unavailable while a robot is performing a task. Different cameras provide
different views of the workspace: wrist cameras capture details near the
robot's gripper, while third-person cameras provide a broader view of the
scene. When a camera becomes unavailable, the remaining views may not provide
the information previously available from that camera. Unlike changes in
lighting or camera pose \citep{pumacay2024colosseum,Fei_2026_CVPR,wang2024vlatest},
these visual interruptions remove a source of information altogether. This
motivates a research question: \textit{How well can VLA models perform
manipulation when some or all of their camera inputs are unavailable?}

\begin{figure}[t]
\centering
\includegraphics[width=\textwidth]{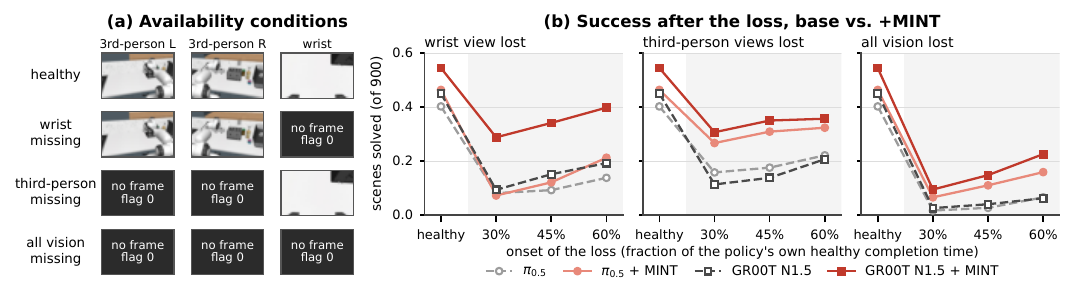}
\caption{\textbf{\mail{}.} (a) Four availability conditions on RoboCasa365;
an interrupted camera delivers no frame and its availability flag is 0.
(b) Scenes solved out of 900 with every camera present (healthy) and after one
visual role is lost at 30, 45 or 60\% of each policy's own completion time:
camera loss removes a large share of both base policies' successes, and
\mint{} recovers part of that loss on every role.}
\label{fig:protocol}
\end{figure}

Existing robustness benchmarks
\citep{pumacay2024colosseum,chen2025robotwin2,Fei_2026_CVPR,wang2024vlatest,hendrycks2019robustness}
primarily examine changes in visual appearance, camera pose, or image quality
while keeping camera streams available. Missing camera inputs pose a
different problem: a policy no longer receives observations from one or more
of its original camera views. The impact may vary across cameras and tasks,
depending on what information remains available from the other views. Yet
existing robustness evaluations provide limited insight into how VLA models
perform under these conditions.

To investigate this problem, we introduce \textbf{\mail{}}, a paired,
closed-loop benchmark for evaluating VLA models with missing camera inputs.
Built on RoboCasa365 \citep{nasiriany2024robocasa,nasiriany2026robocasa365},
it covers 18 manipulation tasks and 900 frozen scenes. Rather than removing
cameras individually, \mail{} evaluates the loss of distinct visual roles:
the wrist view, both third-person views, and all vision. For each policy, we
first identify the scenes it can complete with all cameras available and use
its successful rollouts as references. We then remove the relevant camera
inputs at 30\%--60\% of the policy's own completion time and keep them
unavailable for the remainder of the episode. Missing frames are delivered as
absent rather than replaced by the benchmark with blank or frozen images. By
comparing paired rollouts in the same scenes and measuring success
conditional on intact-camera success, \mail{} quantifies how much capability
a policy retains when visual inputs become unavailable.

Training VLA models with missing camera inputs
\citep{skand2025masked,disdp2025} enables them to continue acting without the
unavailable views, but it cannot recover the visual information those views
would have provided. Predicting missing views with action-conditioned video
models \citep{guo2025ctrlworld,persistworld2026} offers a way to temporarily
supplement the remaining observations, although prediction errors may
accumulate as the robot and scene change. We therefore propose
\textbf{\mint{}}, which combines training for missing camera inputs with
selective visual prediction. During fine-tuning, \mint{} explicitly marks
unavailable cameras and excludes their visual tokens, allowing the policy to
act using the inputs that remain. At inference time, it uses optical-flow
extrapolation for missing wrist views or all-vision
loss, and an action-conditioned world model \citep{persistworld2026} for
missing third-person views. Predictions are retained only while checks based
on available observations and commanded actions are satisfied. Otherwise,
they are withdrawn, and the policy continues with the corresponding inputs
marked as missing.

Experiments on $\pi_{0.5}$ \citep{black2025pi05} and GR00T~N1.5
\citep{nvidia2025groot} reveal that the impact of missing camera inputs
depends on which views are unavailable: losing a single wrist camera can be
more disruptive than losing both third-person cameras. \mint{} improves task
success under missing camera inputs on both models and, on GR00T~N1.5,
outperforms a matched clean fine-tuning control with the same training budget. We further
deploy \mint{} on an AgiBot~G2, where the same
stack runs in real time and the robot completes manipulation tasks in closed
loop with a camera interrupted partway through the task.

\section{Related work}
\label{sec:related}

\paragraph{Robustness Evaluation of Manipulation Policies.}
Generalist suites measure competence with intact observations
\citep{liu2023libero,nasiriany2024robocasa,nasiriany2026robocasa365,mees2022calvin,james2020rlbench,li2023behavior1k,tao2025maniskill3},
and perturbation benchmarks
\citep{pumacay2024colosseum,chen2025robotwin2,Fei_2026_CVPR,wang2024vlatest,li2024simplerenv} change appearance, geometry, language, camera pose or the simulation-to-real gap, with every frame still delivered. 
Recent VLA methods have also improved robustness and generalization through task-aware visual selection and run-time intervention~\citep{xu2026afi,feng2026see}.
Common-corruption suites~\citep{hendrycks2019robustness} degrade image content in the same sense, and a comparative study weighs world action models against VLAs under such perturbations \citep{wamvsvla2026}. 
\mail{} instead evaluates the complete loss of a visual stream during an
ongoing rollout, rather than a change to the content of an available frame. A camera stops delivering frames altogether, at a moment defined by the evaluated policy's own healthy trajectory rather than at a fixed step. The loss is applied per visual role rather than per device, the healthy and interrupted rollouts of a
scene are paired by construction, and the outcome is scored conditional on
healthy success (Table~\ref{tab:peers}).

\paragraph{Learning with Imperfect or Missing Modalities.}
Robust policy learning has been studied under imperfect supervision, including
learning from suboptimal or noisy demonstrations~\citep{wang2023unlabeled, wu2019imitation,wang2021learning,wang2024imitation}.
Closer to our setting, prior work improves robustness to absent sensors through
masked multimodal training, camera dropout, missing-state inference, retrieval,
and availability-aware representations
\citep{skand2025masked,disdp2025,robopanoptes2025,vogtlowell2026sensors,
rl4il2026,wang2023shaspec,kim2026must,xu2025dispro,gao2026ra}.
\mint{}'s fine-tuning stage follows the masked-training line without adding a
component to the policy.


\begin{table}[t]
\centering
\small
\setlength{\tabcolsep}{7pt}
\caption{Comparison of visual-robustness evaluation settings.
\mail{} uniquely targets persistent visual-stream loss during closed-loop execution.}
\label{tab:peers}

\begin{tabular}{@{}lccc@{}}
\toprule
Benchmark &
Visual failure &
Onset &
Post-loss control \\
\midrule

COLOSSEUM~\citep{pumacay2024colosseum}
& Perturbation & -- & -- \\

RoboTwin~2.0~\citep{chen2025robotwin2}
& Perturbation & -- & -- \\

VLATest~\citep{wang2024vlatest}
& Perturbation & -- & -- \\

LIBERO-Plus~\citep{Fei_2026_CVPR}
& Black frame & Start & Closed loop \\

Missing-modality IL~\citep{rl4il2026}
& Camera dropout & Start & Open loop \\

\midrule
\textbf{\mail{} (ours)}
& \textbf{Camera loss}
& \textbf{Mid-episode}
& \textbf{Closed loop} \\

\bottomrule
\end{tabular}
\end{table}

\section{\mail{}: manipulation under visual interruption}
\label{sec:mail}

We introduce \mail{}, a benchmark for evaluating how visual-stream interruptions affect manipulation policies on tasks they can otherwise complete. 

\subsection{Benchmark Construction}

\noindent\textbf{Environment.}
We build \mail{} on RoboCasa365
\citep{nasiriany2026robocasa365},
which provides diverse kitchen environments and everyday
manipulation tasks. We select its 18 Atomic-Seen tasks to
reduce the confounding effect of task generalization and
focus on camera-loss robustness in previously learned tasks.
With 50 fixed scenes per task, MAIL-Bench comprises 900
evaluation scenes.

\noindent\textbf{Availability conditions.}
To make camera-loss conditions comparable across policies with different camera configurations, we group cameras by their visual roles rather than by individual devices. 
The wrist camera forms the \emph{wrist} group: it moves with the gripper and observes the manipulated object at close range. The onboard cameras form the
\emph{third-person} group: they are fixed to the robot base and provide broader views of the arm and workspace. 
Then we consider four availability conditions. \emph{Healthy}
keeps all cameras available and serves as the reference rollout.
\emph{Wrist missing} removes the wrist view, \emph{third-person missing}
removes all onboard views, and \emph{all vision missing} removes all visual
input, leaving only proprioception and the instruction. 
We use \emph{hard missing}: once interrupted, a camera group provides no further frames and its availability flag is set to 0.

\noindent\textbf{Interruption timing.} Camera loss may occur at different points during task execution. To evaluate this variation, we introduce interruptions at 30\%, 45\%, and 60\% of each policy's successful healthy completion time. Defining onset relative to the policy's own execution accounts for differences
in execution speed, rather than imposing the same absolute
control step on every policy. Together, the three camera-loss conditions and three onset times define nine interruption settings, evaluated against a healthy reference.

\begin{figure}[t]
\centering
\includegraphics[width=\textwidth]{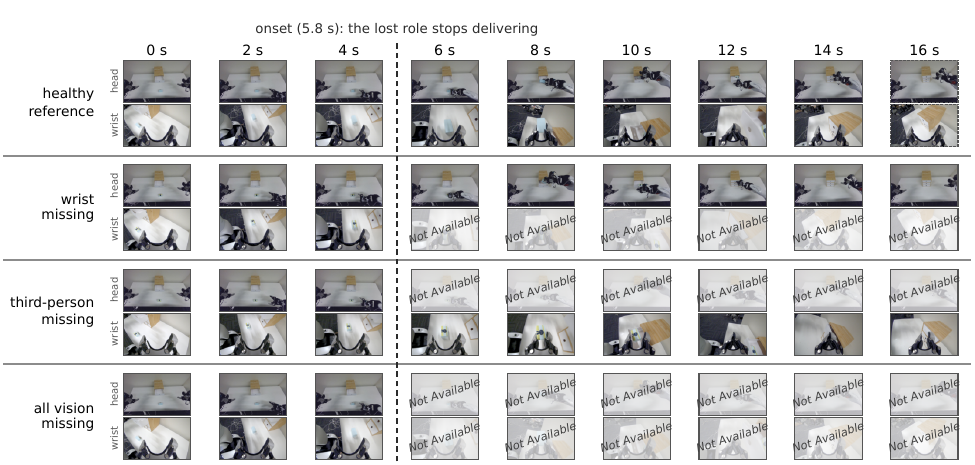}
\caption{
A successful healthy rollout defines the reference completion time
$T_{\mathrm{healthy}}$. The same scene is then replayed with the selected
visual role interrupted at 30\%, 45\%, or 60\% of
$T_{\mathrm{healthy}}$ and kept unavailable thereafter, while unaffected
cameras continue streaming. Shown is an AgiBot~G2 drawer example.}
\label{fig:construction}
\end{figure}

\subsection{Benchmark Evaluation}
\label{sec:contract}

\paragraph{Evaluation protocol.} 
For each scene, we save the episode metadata and a settled simulator state immediately before policy execution. This state serves as the common starting
point for the healthy rollout and all interruption rollouts. For each policy configuration, we first execute a healthy rollout with all cameras available.
If the rollout succeeds, we record its completion step, $T_{\mathrm{healthy}}$; otherwise, the scene is excluded from interruption evaluation.

Each interruption rollout restarts from the same saved initial state as the healthy rollout and uses the same policy random seed. It receives normal camera observations until the designated interruption control step, which, for onset fraction $f\in\{0.30,0.45,0.60\}$, is defined as
\begin{equation}
\label{eq:onset}
T_f =
\min\!\left(
T_{\max}-1,\,
\left\lfloor f\times T_{\mathrm{healthy}}+0.5\right\rfloor
\right),
\end{equation}
where $T_{\max}$ denotes the task horizon. 

Let $\mathcal{R}$ denote the set of visual roles. The policy receives observations only when it is queried, every few control steps; for each role $r\in\mathcal{R}$, let $I_q^r$ denote its camera observation at the $q$-th query and $m_q^r\in\{0,1\}$ indicate whether it is available. When a camera becomes unavailable at $T_f$, the benchmark discards the remaining actions of the current chunk and queries the policy immediately; we denote this query by $q_f$. Before $q_f$, $m_q^r=1$; for an interrupted role, from $q_f$ onward $m_q^r=0$ and the visual input is replaced by $\varnothing$, indicating that no fresh frame is provided.

After the interruption, subsequent frames from the interrupted camera group are withheld from both the policy and any recovery module. Any replacement view must therefore be generated causally from information available at runtime, such as pre-interruption observations, surviving camera streams, and executed actions, without access to future simulator frames or states. Each successful healthy rollout gives rise to nine fault rollouts, covering three camera-loss conditions and three interruption onsets. Deterministic execution settings and checks for pre-interruption divergence are used to reduce variation unrelated to the camera intervention.

\paragraph{Evaluation metrics.} 
We measure robustness only on scenes that the policy can already solve with
all cameras available. For task $j$, let $n_j=50$ be the total number of
scenes, $N^H_j$ the number solved under the healthy condition, and $N^c_j$
the number of those same scenes that remain successful under interruption
condition $c$. We define
\begin{equation}
\label{eq:score}
H_j=\frac{N^{H}_j}{n_j},\qquad
M_{c,j}=\frac{N^{c}_j}{N^{H}_j},\qquad
S_j=\frac{1}{10}\Big(H_j+\sum_{c=1}^{9}M_{c,j}\Big),\qquad
S_{\mathrm{MAIL}}=\frac{1}{18}\sum_{j=1}^{18}S_j .
\end{equation}
Here, $H_j$ measures healthy task success, while $M_{c,j}$ measures how much
of that capability is retained after interruption $c$. Conditioning
$M_{c,j}$ on healthy success prevents robustness from being dominated by a
policy's baseline task competence. For each task, healthy performance and the
nine interruption conditions are weighted equally, and the final
$S_{\mathrm{MAIL}}$ score averages equally over all 18 tasks. We evaluate
every interruption condition even if a policy does not normally rely on the
corresponding camera, since remaining unaffected by its removal is itself
evidence of robustness. If a policy has no healthy success on a task, that
task receives a score of zero.

\section{\mint{}: Acting under Visual Interruption}
\label{sec:method}

\begin{figure}[t]
\centering
\includegraphics[width=\textwidth]{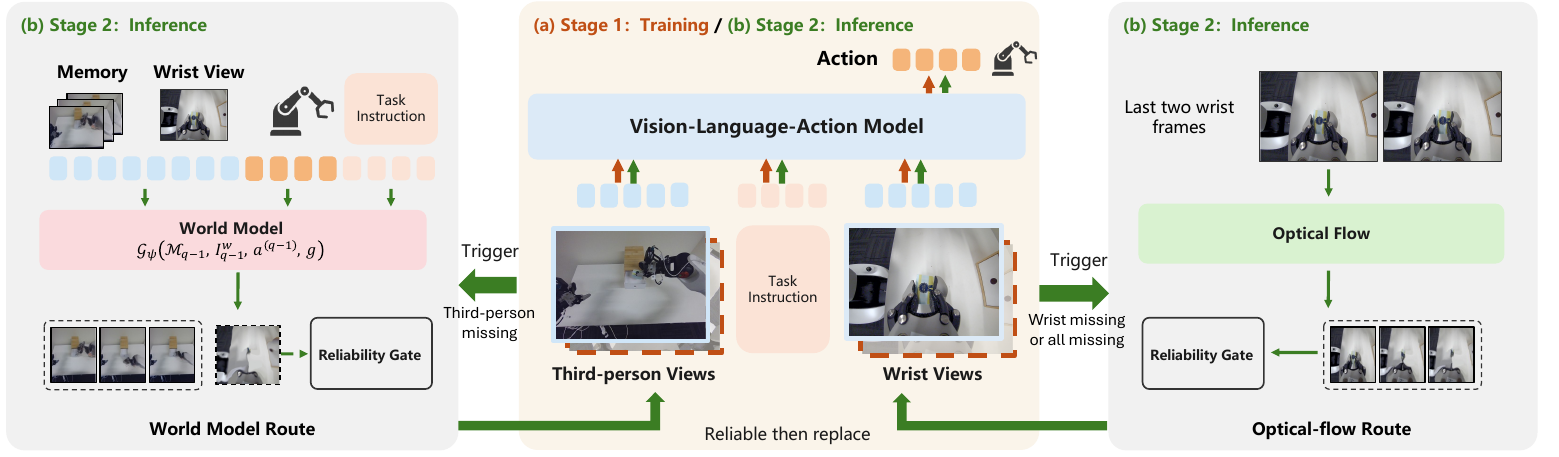}
\vskip 0.1in
\caption{\textbf{\mint{} overview.}
\mint{} separates robustness to missing visual inputs from visual prediction.
The policy is first trained to act under variable camera availability, then
selectively receives predicted views at inference time when a camera is
interrupted. Unreliable predictions are withdrawn, and the policy falls back
to acting from the observations that remain.}
\label{fig:overview}
\end{figure}

A robust VLA policy should remain functional when a camera becomes
unavailable, without relying on the missing view being reconstructed. At the
same time, when that view can be predicted reliably, the prediction may
temporarily recover useful information that is no longer directly observed.
\mint{} separates these two capabilities: learning with missing visual inputs
teaches the policy to act from the observations that remain, while selective
visual prediction supplements a missing view only when its prediction is
considered reliable.

At the $q$-th query, given visual observations $\{I_q^r\}_{r\in\mathcal{R}}$,
proprioceptive state $s_q$, and task instruction $g$, the policy predicts an
action chunk of $h$ actions
\begin{equation}
\label{eq:vla}
a^{(q)} \sim
\pi_\theta\!\left(
\cdot \mid
\{I_q^r\}_{r\in\mathcal{R}},\, s_q,\, g
\right),
\end{equation}
where $h$ denotes the action horizon; only the first few actions are executed
before the next query. Under visual interruption, one or more
roles become unavailable during execution. \mint{} first adapts the policy to
such changes in observation availability and then, at inference time,
selectively supplies predicted observations for missing roles when reliable
estimates are available.

\subsection{Learning with Missing Visual Inputs}
\label{sec:policy}

We first train the policy to predict actions when only a subset of its usual
visual observations is available. When $m^r=0$, the corresponding
view is marked as missing and its visual tokens are excluded from attention,
matching the hard-missing semantics of \mail{}.

Each training sample is a demonstration frame with observations
$\{I^r\}_{r\in\mathcal{R}}$, state $s$, instruction $g$, and the next $h$
demonstrated actions $a^{\ast}$. An availability pattern $m\sim\mathcal{C}$,
sampled over the demonstration episode, determines which roles are missing at
that frame, and $\tilde I^r$ denotes the input of role $r$ under $m$. Using
the backbone's original action objective $\ell$, we optimize
\begin{equation}
\label{eq:train}
\mathcal{L}(\theta)=
\mathbb{E}_{(\{I^r\},s,g,a^{\ast})\sim\mathcal{D},\,m\sim\mathcal{C}}
\left[
\ell\!\left(
\pi_\theta\!\left(
\cdot \mid
\{(\tilde I^r,m^r)\}_{r\in\mathcal{R}},\,s,\,g
\right),
a^{\ast}
\right)
\right],
\end{equation}
where $\mathcal{C}$ is a distribution over temporal camera-availability
patterns. It includes healthy, wrist-missing, third-person-missing, and
all-vision-missing conditions, with both persistent and intermittent
interruptions.

The policy architecture, optimization schedule, and action objective are
otherwise unchanged. The resulting policy can therefore continue acting from
the observations that remain when no reliable estimate of a missing view is
available.

\subsection{Predicting Missing Visual Inputs}
\label{sec:prediction}

Learning with missing visual inputs enables the policy to continue acting when
a camera role is unavailable, but it cannot recover information that is no
longer observed. At inference time, \mint{} therefore constructs a candidate
prediction for a missing role and supplies it to the policy only while the
prediction remains reliable.

Let $\hat I_q^r$ denote a candidate prediction for visual role $r$ at query
$q$, and let $c_q^r\in\{0,1\}$ indicate whether that prediction is currently
accepted. The observation presented to the policy is
\begin{equation}
\label{eq:routing}
(x_q^r,\bar m_q^r)=
\begin{cases}
(I_q^r,1),
& m_q^r=1,\\
(\hat I_q^r,1),
& m_q^r=0 \ \wedge\ c_q^r=1,\\
(\varnothing,0),
& m_q^r=0 \ \wedge\ c_q^r=0,
\end{cases}
\end{equation}
where $\bar m_q^r$ is the availability state seen by the policy. Thus, a
missing role is either temporarily replaced by a predicted observation or
left explicitly missing, in which case the policy falls back to the behavior
learned in Section~\ref{sec:policy}.

\noindent\textbf{Optical-flow prediction.} The prediction mechanism depends on the visual role that becomes unavailable.
For wrist interruption, and for all-vision interruption where no surviving
camera is available to validate a generative rollout, we use short-horizon
optical-flow extrapolation. From the last two available frames, we estimate
\begin{equation}
\label{eq:flow}
F^r =
\operatorname{Flow}\!\left(
I_{q_f-2}^r,I_{q_f-1}^r
\right),
\qquad
\hat I_q^r =
\operatorname{Warp}\!\left(
I_{q_f-1}^r,F^r,q-q_f+1
\right),
\quad q\ge q_f ,
\end{equation}
where $\operatorname{Flow}$ is Farneb\"ack optical flow
\citep{farneback2003} and $\operatorname{Warp}$ denotes short-horizon
extrapolation from the last observed frame. This route is appropriate for
wrist views, whose short-term visual dynamics are dominated by camera motion,
and provides a conservative prediction when all visual streams are lost.

\noindent\textbf{World-model prediction.} Third-person interruption requires modeling changes in both robot and object configuration that cannot be captured reliably by warping the last frame. We therefore use an action-conditioned world model. At each policy query $q$,
the model predicts the missing third-person views together with an auxiliary
wrist view:
\begin{equation}
\label{eq:wm}
(\hat I_q^{L},\hat I_q^{R},\hat I_q^{w})
=
\mathcal{G}_{\psi}\!\left(
\mathcal{M}_{q-1},
I_{q-1}^{w},
a^{(q-1)},
g
\right),
\end{equation}
where $\mathcal{M}_{q-1}$ is the model's rollout memory, initialized at the
interruption from the last available observations and updated with its own
predictions, $I_{q-1}^{w}$ is the
surviving wrist observation, $a^{(q-1)}$ is the committed action chunk, and
$g$ is the task instruction. We instantiate $\mathcal{G}_{\psi}$ with
PersistWorld~\citep{persistworld2026}, adapted to the task domain with a
low-rank fine-tune.

Only $\hat I_q^{L}$ and $\hat I_q^{R}$ are supplied to the policy. The
auxiliary wrist prediction $\hat I_q^{w}$ serves a different purpose:
because the real wrist view remains observable, its agreement with
$\hat I_q^{w}$ provides an online consistency signal for the otherwise
unobserved third-person rollout. Section~\ref{sec:reliability} describes how
this signal, together with route-specific checks, determines $c_q^r$.

\subsection{Reliability-Gated Fallback}
\label{sec:reliability}

Predicted observations are intended only as temporary substitutes for missing views. Because prediction errors can accumulate after interruption, \mint{} continuously monitors each prediction route using a set of route-specific reliability checks. A route remains active only while all of its checks are satisfied. Once rejected, its prediction is withdrawn for the remainder of the episode and the corresponding visual role falls back to the missing-input state learned in Section~\ref{sec:policy}.

For a missing visual role $r$, let $\{z_{q,k}^r\}_{k\in\mathcal{K}_r}$ denote its reliability diagnostics at query $q$, where $\mathcal{K}_r$ indexes the route-specific checks. We define route reliability as
\begin{equation}
\label{eq:reliability}
\rho_q^r =
\prod_{k\in\mathcal{K}_r}
\mathbf{1}
\left[
z_{q,k}^r \in \mathcal{A}_{r,k}
\right],
\end{equation}
where $\mathcal{A}_{r,k}$ denotes the admissible range of the $k$-th
reliability diagnostic (Table~\ref{tab:mint-constants}). Thus, $\rho_q^r=1$ only when all route-specific
checks are satisfied.

The acceptance state introduced in Eq.~\ref{eq:routing} is updated as
\begin{equation}
\label{eq:gate}
c_q^r =
c_{q-1}^r \wedge \rho_q^r,
\qquad
c_{q_f}^r = 1 .
\end{equation}
This recursive update makes rejection irreversible within an episode: once $c_q^r=0$, predicted observations for that role are no longer supplied to the policy.

\paragraph{Optical-flow route.} The optical-flow route uses two reliability diagnostics: consistency of the propagated view and accumulated robot motion. Both must remain within their
respective admissible ranges in Eq.~\ref{eq:reliability}. If either check fails, $\rho_q^r=0$, causing the route to be rejected through Eq.~\ref{eq:gate}.

\paragraph{World-model route.}
The world-model route uses four reliability diagnostics: wrist consistency, visual change, model-distribution deviation, and rollout horizon. Wrist consistency compares an auxiliary predicted wrist observation $\hat I_q^w$ with the corresponding real wrist observation $I_q^w$:
\begin{equation}
\label{eq:wrist_consistency}
d_q^w =
D\!\left(
\hat I_q^w,\,
I_q^w
\right),
\end{equation}
where $D(\cdot,\cdot)$ measures normalized visual disagreement. The resulting $d_q^w$, together with the other three diagnostics, forms the set of route-specific checks in Eq.~\ref{eq:reliability}. If any diagnostic falls outside its admissible range, $\rho_q^r=0$ and the route is rejected through Eq.~\ref{eq:gate}.

Once a prediction route is withdrawn, control continues through the policy's learned missing-input capability rather than relying on continued visual reconstruction.

\section{Experiments}
\label{sec:exp}

We evaluate \mint{} across two VLA backbones on \mail{}, use matched controls to disentangle the contributions of learning with missing visual inputs and selective visual prediction, and further examine closed-loop deployment on a physical AgiBot-G2 robot.

\subsection{Experimental Setup}
\label{sec:mail-eval}

\noindent\textbf{Models.}
We evaluate the RoboCasa365 Human300-trained checkpoints of
$\pi_{0.5}$~\citep{black2025pi05} and GR00T~N1.5~\citep{nvidia2025groot},
together with their respective \mint{} variants. The base models use their original policy implementations, and \mint{} is applied without modifying the underlying VLA architecture. 
All policies are evaluated on the frozen \mail{} suite of 18 Atomic-Seen tasks and 900 scenes described in Section~\ref{sec:mail}. We divide 18 tasks into 3 categories, which are 5 \textit{pick-and-place} tasks, 6 \textit{small-appliance} tasks, and 7 \textit{fixture and navigation} tasks. We report the results of each category in Table \ref{tab:mail-grid-categories}. The complete task list and results are reported in Appendix~\ref{app:mail-tables}.

\noindent\textbf{Training.}
Both \mint{} policies are continued for \textit{30k} steps using the original action objective and optimization setup of their original checkpoints. Training uses a temporal availability curriculum over healthy, wrist-missing,
third-person-missing and all-vision-missing observations with probabilities
$\{0.60,0.17,0.17,0.06\}$. For fault episodes, persistent interruptions are
sampled with probability $0.7$, with onset uniformly drawn from
$[0.3,0.7]$ of the episode. The remaining cases contain short intermittent
interruptions. Unavailable visual tokens are masked from policy computation
rather than replaced by informative image content.

\noindent\textbf{Ablation Setup.}
For GR00T~N1.5, we construct two matched variants to isolate the two main
components of \mint{}. \textbf{(1)} \emph{Clean FT + prediction} is fine-tuned for the
same 30k steps using the same training data as \mint{}, but without missing-view examples during training. At inference time, it uses the same prediction mechanism as \mint{}. \textbf{(2)} \emph{Missing-view FT} uses the same missing-view-trained checkpoint as \mint{} and the same serving cadence, but disables generated observations so that interrupted views remain missing. Thus,
\emph{Clean FT + prediction} versus \mint{} isolates the contribution of
missing-view training under matched inference, while \emph{Missing-view FT}
versus \mint{} isolates the contribution of selective visual prediction
with the policy weights held fixed.
More implementation details are provided in Appendix~\ref{app:mint-details}.

\begin{table}[t]
\caption{\textbf{MAIL-Bench results.}
Healthy denotes success rate with all cameras available, while W/T/B report
scores under wrist, third-person, and all-vision interruption
at 30/45/60\% of completion time. AVG is the
\mail{} score and $\Delta$ its improvement over the corresponding base
checkpoint. \textbf{Bold} marks the best result within each backbone and
task category. 
}
\label{tab:mail-grid-categories}
\centering\scriptsize
\setlength{\tabcolsep}{3pt}
\setlength{\fboxsep}{1pt}
\resizebox{\textwidth}{!}{
\begin{tabular}{l l c ccc ccc ccc c c}
\toprule
\multirow{2}{*}{Method}
& \multirow{2}{*}{Tasks}
& \multirow{2}{*}{Healthy}
& \multicolumn{3}{c}{Wrist missing}
& \multicolumn{3}{c}{Third-person missing}
& \multicolumn{3}{c}{All vision missing}
& \multirow{2}{*}{AVG}
& \multirow{2}{*}{$\Delta$} \\
\cmidrule(lr){4-6}
\cmidrule(lr){7-9}
\cmidrule(lr){10-12}
& & & W30 & W45 & W60 & T30 & T45 & T60 & B30 & B45 & B60 & & \\
\midrule

\multicolumn{14}{l}{\textit{GR00T N1.5}~\citep{nvidia2025groot}} \\

base model & pick-and-place
& \textbf{0.732}
& 0.163 & 0.276 & 0.341
& 0.165 & 0.224 & 0.424
& 0.000 & 0.000 & 0.010
& 0.234 & - \\

& small appliances
& 0.290
& 0.153 & 0.277 & 0.398
& 0.245 & 0.296 & 0.403
& 0.110 & 0.170 & 0.226
& 0.257 & - \\

& fixtures \& navigation
& 0.386
& 0.178 & 0.253 & 0.385
& 0.308 & 0.319 & 0.398
& 0.029 & 0.084 & 0.119
& 0.246 & - \\

\rowcolor{black!8}
\textbf{+\mint{}} & pick-and-place
& 0.724
& \textbf{0.423} & \textbf{0.561} & \textbf{0.673}
& \textbf{0.622} & \textbf{0.674} & \textbf{0.647}
& \textbf{0.053} & \textbf{0.091} & \textbf{0.166}
& \textbf{0.463} & +0.229 \\

\rowcolor{black!8}
& small appliances
& \textbf{0.450}
& \textbf{0.493} & \textbf{0.536} & \textbf{0.631}
& \textbf{0.602} & \textbf{0.628} & \textbf{0.573}
& \textbf{0.307} & \textbf{0.419} & \textbf{0.528}
& \textbf{0.517} & +0.260 \\

\rowcolor{black!8}
& fixtures \& navigation
& \textbf{0.497}
& \textbf{0.450} & \textbf{0.507} & \textbf{0.649}
& \textbf{0.335} & \textbf{0.488} & \textbf{0.566}
& \textbf{0.076} & \textbf{0.197} & \textbf{0.378}
& \textbf{0.414} & +0.168 \\

\midrule

\multicolumn{14}{l}{\textit{$\pi_{0.5}$}~\citep{black2025pi05}} \\

base model & pick-and-place
& 0.636
& \textbf{0.059} & 0.098 & 0.200
& 0.239 & 0.245 & 0.375
& 0.000 & 0.014 & 0.089
& 0.196 & - \\

& small appliances
& 0.257
& 0.270 & 0.264 & 0.342
& 0.291 & 0.343 & 0.413
& 0.065 & 0.095 & 0.156
& 0.250 & - \\

& fixtures \& navigation
& 0.360
& \textbf{0.151} & \textbf{0.266} & 0.355
& \textbf{0.453} & 0.469 & \textbf{0.602}
& 0.043 & \textbf{0.129} & 0.254
& 0.308 & - \\

\rowcolor{black!8}
\textbf{+\mint{}} & pick-and-place
& \textbf{0.688}
& 0.027 & \textbf{0.111} & \textbf{0.342}
& \textbf{0.574} & \textbf{0.735} & \textbf{0.713}
& \textbf{0.025} & \textbf{0.090} & \textbf{0.196}
& \textbf{0.350} & +0.154 \\

\rowcolor{black!8}
& small appliances
& \textbf{0.333}
& \textbf{0.291} & \textbf{0.426} & \textbf{0.350}
& \textbf{0.513} & \textbf{0.372} & \textbf{0.480}
& \textbf{0.218} & \textbf{0.337} & \textbf{0.370}
& \textbf{0.369} & +0.119 \\

\rowcolor{black!8}
& fixtures \& navigation
& \textbf{0.414}
& 0.131 & 0.261 & \textbf{0.528}
& 0.428 & \textbf{0.498} & 0.546
& \textbf{0.093} & 0.128 & \textbf{0.278}
& \textbf{0.331} & +0.023 \\

\bottomrule
\end{tabular}}

\end{table}

\subsection{Benchmark Results}

\noindent\textbf{Overall Results.}
As demonstrated in Table \ref{tab:mail-grid-categories}, \mint{} improves the performance of VLA base model even with healthy observations. On GR00T~N1.5, for example, healthy success on the \textit{small-appliance} tasks increases from 0.290 to 0.450. Under healthy observation setting, we only have model with availability finetuning and there is no visual generation process for the missing view. This suggests that learning under partial visual observations does not introduce a trade-off with normal execution and may reduce over-reliance on view-specific superficial cues.
Once visual inputs are interrupted, we observe more clear improvement. Across 9 interruption conditions, \mint{} outperforms the GR00T~N1.5 and $\pi_{0.5}$ base model in almost all cases. The gain spans wrist, third-person, and all-vision loss and persists across early and late interruption onsets, indicating that the benefit is not tied to a particular failure pattern. The consistent improvement across two VLA backbones (\textit{i.e.}, GR00T N1.5 and $\pi_{0.5}$) further suggests that \mint{} is not specific to a single policy architecture.

\begin{wrapfigure}{r}{0.55\textwidth}
\vspace{-12pt}
\centering
\includegraphics[width=0.56\textwidth]{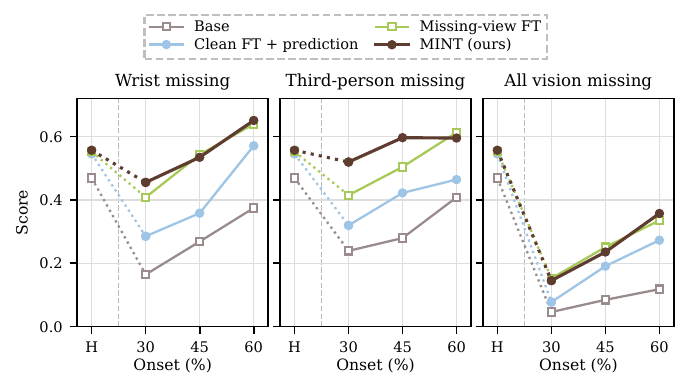}
\vspace{-16pt}
\caption{\textbf{Ablation on GR00T~N1.5.}
Scores averaged over the three task categories at each interruption onset.
``H'' marks the healthy-input score of each variant.}
\label{fig:ablation}
\end{wrapfigure}
\noindent\textbf{Ablation Study.}
\mint{} consists of two stages: missing-view training and selective visual
prediction at inference and we ablate each on GR00T~N1.5. Removing either stage degrades score, and Fig.~\ref{fig:ablation} shows that the two stages cover different failure regimes. Missing-view training governs how far the policy falls when a view is lost early: without it, scores at the
30\% onset drop to 0.29/0.32 under wrist/third-person loss, whereas late
interruptions (W60) are largely recovered by prediction alone. Prediction, in
turn, determines how much of the lost view can be recovered when another view
remains: it is most effective under early third-person interruption
($+0.105$ at T30), where the global view can still be inferred from the wrist
camera, and its gain shrinks toward the 60\% onset as less of the task
remains. Under all-vision loss the two curves coincide, since the only prediction available there is short-horizon flow extrapolation of the last frames. 

\begin{figure}[t]
\centering
\includegraphics[width=\textwidth]{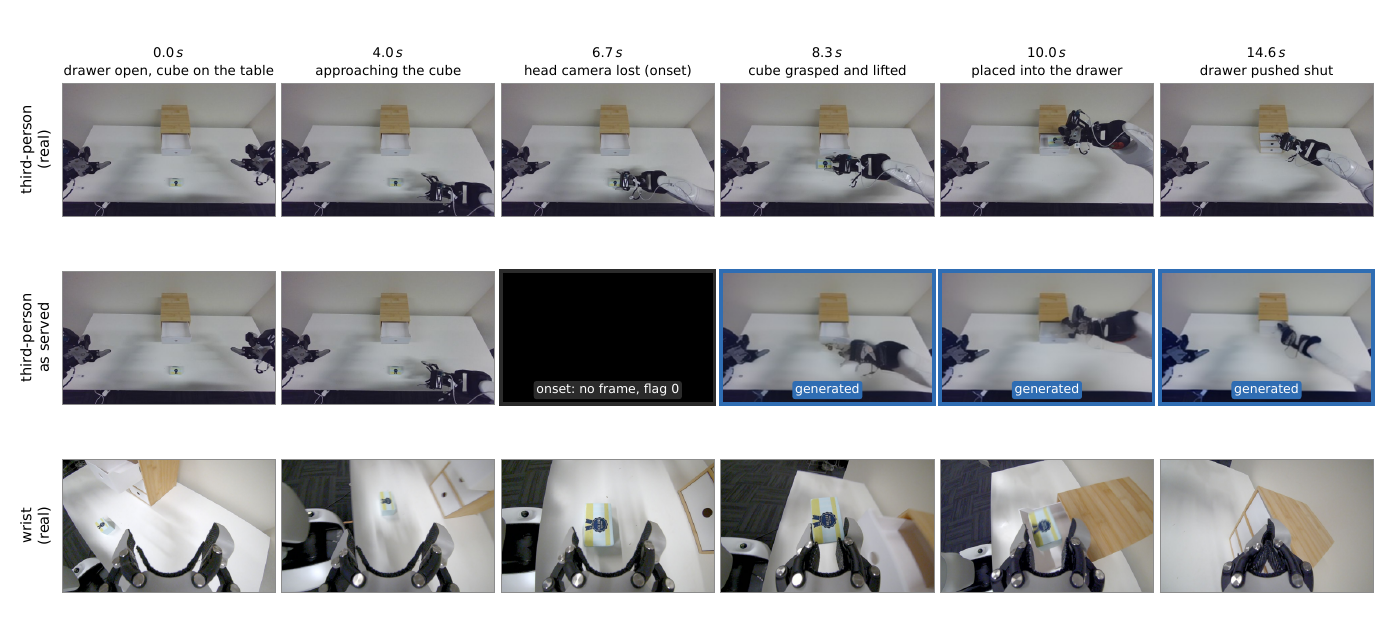}
\caption{Episode on `Cube to Drawer' task with the third-person camera interrupted at step 50. The policy completes grasping, transport, release and drawer closure. \textbf{Top:} real third-person stream, hidden from the policy after interruption. \textbf{Middle:} the third-person view stream and prediction by the world model. \textbf{Bottom:} the wrist stream, which stays available.}
\label{fig:g2-drawer-world}
\end{figure}
\begin{figure}[t]
\centering
\includegraphics[width=\textwidth]{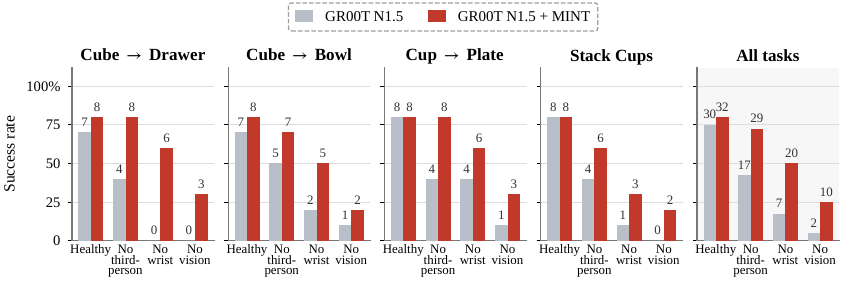}
\caption{\textbf{Real robot experiment on the AgiBot~G2.}
Success rate of GR00T~N1.5 and \mint{} over 10 trials per task and condition. Numbers above the bars are successful trials. 
}
\label{fig:g2-bars}
\end{figure}

\subsection{Real Robot Experiment}
\label{sec:physical}

We next ask whether the same observation-interface design transfers to a physical robot without changing the underlying VLA architecture. We deploy \mint{} on an AgiBot~G2 using an availability-aware GR00T~N1.5 policy fine-tuned on 203 successful teleoperated episodes spanning four tabletop manipulation tasks. The G2 head camera provides the third-person view and the right wrist camera the wrist view. On all four tasks, we compare GR00T~N1.5 with \mint{} under healthy execution and under third-person, wrist and all-camera interruption. The results are shown in Figure~\ref{fig:g2-bars}. We can observe that with \mint{}, GR00T N1.5 outperforms the base model in all missing-input settings and has no performance degradation under healthy setting.

Figure~\ref{fig:g2-drawer-world} shows a drawer episode in which the third-person camera is interrupted mid-episode and the world-model route is activated. The healthy reference episode is shown in Appendix~\ref{app:g2} (Figure~\ref{fig:g2-drawer}). After interruption the policy no longer receives real third-person camera frames. From the next query the world model supplies the missing view and the robot completes the task.

\section{Conclusion}
\label{sec:conclusion}
In this paper, we propose a new benchmark \mail{}, which provides a systematic
evaluation of this problem and shows that interruption severity depends
strongly on the visual role that is lost. \mint{} improves robustness by
combining learning under missing visual inputs with selective visual
prediction. Across two VLA backbones, it improves interruption robustness
without modifying the policy architecture, and the same interface
transfers to real-robot deployment.\\
\noindent\textbf{Future Work.}
\mail{} currently covers one embodiment and 18 RoboCasa365 Atomic-Seen tasks,
and broader validation across benchmarks and robot platforms remains future
work. In addition, \mint{}'s world-model route introduces extra inference and
memory overhead, motivating us to develop more efficient future prediction methods.

{\sloppy\raggedright
\bibliography{refs}

@inproceedings{
feng2026see,
title={See What Matters: Differentiable Grid Sample Pruning for Generalizable Vision-Language-Action Model},
author={Yixu Feng and Zinan Zhao and Yanxiang Ma and Chenghao Xia and Chengbin Du and Yunke Wang and Chang Xu},
booktitle={Forty-third International Conference on Machine Learning},
year={2026}
}

@inproceedings{wang2023unlabeled,
  title={Unlabeled imperfect demonstrations in adversarial imitation learning},
  author={Wang, Yunke and Du, Bo and Xu, Chang},
  booktitle={Proceedings of the AAAI conference on artificial intelligence},
  volume={37},
  number={8},
  pages={10262--10270},
  year={2023}
}

@inproceedings{wu2019imitation,
  title={Imitation learning from imperfect demonstration},
  author={Wu, Yueh-Hua and Charoenphakdee, Nontawat and Bao, Han and Tangkaratt, Voot and Sugiyama, Masashi},
  booktitle={International Conference on Machine Learning},
  pages={6818--6827},
  year={2019},
  organization={PMLR}
}

@inproceedings{wang2021learning,
  title={Learning to weight imperfect demonstrations},
  author={Wang, Yunke and Xu, Chang and Du, Bo and Lee, Honglak},
  booktitle={International conference on machine learning},
  pages={10961--10970},
  year={2021},
  organization={PMLR}
}

@inproceedings{wang2024imitation,
author = {Wang, Yunke and Dong, Minjing and Zhao, Yukun and Du, Bo and Xu, Chang},
title = {Imitation learning from purified demonstrations},
year = {2024},
publisher = {JMLR.org},
booktitle = {Proceedings of the 41st International Conference on Machine Learning},
articleno = {2059},
numpages = {19},
location = {Vienna, Austria},
series = {ICML'24}
}

@article{gao2026ra,
  title={RA-SOD: Reliability-Aware RGB-T Salient Object Detection under Modality Degradation},
  author={Gao, Hongbo and Li, Zhengyu and Nie, Xueru and Zhu, Dihao and Zhao, Lijun and Wang, Yunke and Xu, Chang},
  journal={arXiv preprint arXiv:2609.12622},
  year={2026}
}

@inproceedings{xu2026afi,
  title     = {Affordance Field Intervention: Enabling VLAs to Escape Memory Traps in Robotic Manipulation},
  author    = {Xu, Siyu and Wang, Zijian and Wang, Yunke and Xia, Chenghao and Huang, Tao and Xu, Chang},
  booktitle = {Proceedings of the IEEE/CVF Conference on Computer Vision and Pattern Recognition (CVPR)},
  year      = {2026}
}

@inproceedings{liu2023libero,
  title={{LIBERO}: Benchmarking Knowledge Transfer for Lifelong Robot Learning},
  author={Liu, Bo and Zhu, Yifeng and Gao, Chongkai and Feng, Yihao and Liu, Qiang and Zhu, Yuke and Stone, Peter},
  booktitle={Advances in Neural Information Processing Systems 36 (NeurIPS 2023) Datasets and Benchmarks Track},
  volume={36},
  year={2023}
}

@inproceedings{nasiriany2024robocasa,
  title={{RoboCasa}: Large-Scale Simulation of Household Tasks for Generalist Robots},
  author={Nasiriany, Soroush and Maddukuri, Abhiram and Zhang, Lance and Parikh, Adeet and Lo, Aaron and Joshi, Abhishek and Mandlekar, Ajay and Zhu, Yuke},
  booktitle={Robotics: Science and Systems (RSS)},
  year={2024},
  doi={10.15607/RSS.2024.XX.050}
}

@inproceedings{nasiriany2026robocasa365,
  title={{RoboCasa365}: A Large-Scale Simulation Framework for Training and Benchmarking Generalist Robots},
  author={Nasiriany, Soroush and Nasiriany, Sepehr and Maddukuri, Abhiram and Zhu, Yuke},
  booktitle={International Conference on Learning Representations (ICLR)},
  year={2026}
}

@article{mees2022calvin,
  title={{CALVIN}: A Benchmark for Language-Conditioned Policy Learning for Long-Horizon Robot Manipulation Tasks},
  author={Mees, Oier and Hermann, Lukas and Rosete-Beas, Erick and Burgard, Wolfram},
  journal={IEEE Robotics and Automation Letters (RA-L)},
  volume={7},
  number={3},
  pages={7327--7334},
  year={2022},
  doi={10.1109/LRA.2022.3180108}
}

@article{james2020rlbench,
  title={{RLBench}: The Robot Learning Benchmark \& Learning Environment},
  author={James, Stephen and Ma, Zicong and Arrojo, David Rovick and Davison, Andrew J.},
  journal={IEEE Robotics and Automation Letters (RA-L)},
  volume={5},
  number={2},
  pages={3019--3026},
  year={2020},
  doi={10.1109/LRA.2020.2974707}
}

@inproceedings{pumacay2024colosseum,
  title={{THE COLOSSEUM}: A Benchmark for Evaluating Generalization for Robotic Manipulation},
  author={Pumacay, Wilbert and Singh, Ishika and Duan, Jiafei and Krishna, Ranjay and Thomason, Jesse and Fox, Dieter},
  booktitle={Robotics: Science and Systems (RSS)},
  year={2024},
  doi={10.15607/RSS.2024.XX.133}
}

@inproceedings{li2024simplerenv,
  title={Evaluating Real-World Robot Manipulation Policies in Simulation},
  author={Li, Xuanlin and Hsu, Kyle and Gu, Jiayuan and Mees, Oier and Pertsch, Karl and Walke, Homer Rich and Fu, Chuyuan and Lunawat, Ishikaa and Sieh, Isabel and Kirmani, Sean and Levine, Sergey and Wu, Jiajun and Finn, Chelsea and Su, Hao and Vuong, Quan and Xiao, Ted},
  booktitle={Proceedings of the 8th Conference on Robot Learning (CoRL)},
  series={Proceedings of Machine Learning Research},
  volume={270},
  pages={3705--3728},
  year={2025},
  publisher={PMLR}
}

@inproceedings{chen2025robotwin2,
  title={{RoboTwin 2.0}: A Scalable Data Generator and Benchmark with Strong Domain Randomization for Robust Bimanual Robotic Manipulation},
  author={Chen, Tianxing and Chen, Zanxin and Chen, Baijun and Cai, Zijian and Liu, Yibin and Li, Zixuan and Liang, Qiwei and Lin, Xianliang and Ge, Yiheng and Gu, Zhenyu and Deng, Weiliang and Guo, Yubin and Nian, Tian and Xie, Xuanbing and Chen, Qiangyu and Su, Kailun and Xu, Tianling and Liu, Guodong and Hu, Mengkang and Gao, Huan-ang and Wang, Kaixuan and Liang, Zhixuan and Qin, Yusen and Yang, Xiaokang and Luo, Ping and Mu, Yao},
  booktitle={International Conference on Machine Learning (ICML)},
  year={2026}
}

@article{wang2024vlatest,
  title={{VLATest}: Testing and Evaluating Vision-Language-Action Models for Robotic Manipulation},
  author={Wang, Zhijie and Zhou, Zhehua and Song, Jiayang and Huang, Yuheng and Shu, Zhan and Ma, Lei},
  journal={Proceedings of the ACM on Software Engineering (FSE)},
  volume={2},
  number={FSE},
  pages={1615--1638},
  year={2025},
  doi={10.1145/3729343}
}

@inproceedings{Fei_2026_CVPR,
  title={{LIBERO-Plus}: A Progressive Robustness Benchmark for Visual-Language-Action Models},
  author={Fei, Senyu and Wang, Siyin and Shi, Junhao and Dai, Zihao and Cai, Jikun and Qian, Pengfang and Ji, Li and He, Xinzhe and Zhang, Shiduo and Fei, Zhaoye and Fu, Jinlan and Gong, Jingjing and Qiu, Xipeng},
  booktitle={Proceedings of the IEEE/CVF Conference on Computer Vision and Pattern Recognition (CVPR)},
  pages={38574--38583},
  year={2026}
}

@inproceedings{tao2025maniskill3,
  title={Demonstrating {GPU} Parallelized Robot Simulation and Rendering for Generalizable Embodied {AI} with {ManiSkill3}},
  author={Tao, Stone and Xiang, Fanbo and Shukla, Arth and Qin, Yuzhe and Hinrichsen, Xander and Yuan, Xiaodi and Bao, Chen and Lin, Xinsong and Liu, Yulin and Chan, Tse-Kai and Gao, Yuan and Li, Xuanlin and Mu, Tongzhou and Xiao, Nan and Gurha, Arnav and Rajesh, Viswesh Nagaswamy and Choi, Yong Woo and Chen, Yen-Ru and Huang, Zhiao and Calandra, Roberto and Chen, Rui and Luo, Shan and Su, Hao},
  booktitle={Robotics: Science and Systems (RSS)},
  year={2025},
  doi={10.15607/RSS.2025.XXI.021}
}

@inproceedings{li2023behavior1k,
  title={{BEHAVIOR-1K}: A Benchmark for Embodied {AI} with 1,000 Everyday Activities and Realistic Simulation},
  author={Li, Chengshu and Zhang, Ruohan and Wong, Josiah and Gokmen, Cem and Srivastava, Sanjana and Mart{\'\i}n-Mart{\'\i}n, Roberto and Wang, Chen and Levine, Gabrael and Lingelbach, Michael and Sun, Jiankai and others},
  booktitle={Proceedings of the 6th Conference on Robot Learning (CoRL)},
  series={Proceedings of Machine Learning Research},
  volume={205},
  pages={80--93},
  year={2023},
  publisher={PMLR}
}

@inproceedings{hendrycks2019robustness,
  title={Benchmarking Neural Network Robustness to Common Corruptions and Perturbations},
  author={Hendrycks, Dan and Dietterich, Thomas},
  booktitle={International Conference on Learning Representations (ICLR)},
  year={2019}
}

@article{wamvsvla2026,
  title={Do World Action Models Generalize Better than {VLAs}? {A} Robustness Study},
  author={Zhang, Zhanguang and Li, Zhiyuan and Rahmati, Behnam and Yang, Rui Heng and Ma, Yintao and Rasouli, Amir and Pakdamansavoji, Sajjad and Wu, Yangzheng and Zhang, Lingfeng and Cao, Tongtong and Wen, Feng and Wang, Xinyu and Quan, Xingyue and Zhang, Yingxue},
  journal={arXiv preprint arXiv:2603.22078},
  year={2026}
}

@inproceedings{black2025pi05,
  title={$\pi_{0.5}$: a Vision-Language-Action Model with Open-World Generalization},
  author={Black, Kevin and Brown, Noah and Darpinian, James and Dhabalia, Karan and Driess, Danny and Esmail, Adnan and Equi, Michael Robert and Finn, Chelsea and Fusai, Niccolo and Galliker, Manuel Y. and Ghosh, Dibya and Groom, Lachy and Hausman, Karol and Ichter, Brian and Jakubczak, Szymon and Jones, Tim and Ke, Liyiming and LeBlanc, Devin and Levine, Sergey and Li-Bell, Adrian and Mothukuri, Mohith and Nair, Suraj and Pertsch, Karl and Ren, Allen Z. and Shi, Lucy Xiaoyang and Smith, Laura and Springenberg, Jost Tobias and Stachowicz, Kyle and Tanner, James and Vuong, Quan and Walke, Homer and Walling, Anna and Wang, Haohuan and Yu, Lili and Zhilinsky, Ury},
  booktitle={Proceedings of the 9th Conference on Robot Learning (CoRL)},
  series={Proceedings of Machine Learning Research},
  volume={305},
  pages={17--40},
  year={2025},
  publisher={PMLR}
}

@inproceedings{black2024pi0,
  title={$\pi_0$: A Vision-Language-Action Flow Model for General Robot Control},
  author={Black, Kevin and Brown, Noah and Driess, Danny and Esmail, Adnan and Equi, Michael and Finn, Chelsea and Fusai, Niccolo and Groom, Lachy and Hausman, Karol and Ichter, Brian and others},
  booktitle={Robotics: Science and Systems (RSS)},
  year={2025},
  doi={10.15607/RSS.2025.XXI.010}
}

@article{nvidia2025groot,
  title={{GR00T N1}: An Open Foundation Model for Generalist Humanoid Robots},
  author={{NVIDIA} and Bjorck, Johan and Casta{\~n}eda, Fernando and Cherniadev, Nikita and Da, Xingye and Ding, Runyu and Fan, Linxi and Fang, Yu and Fox, Dieter and Hu, Fengyuan and others},
  journal={arXiv preprint arXiv:2503.14734},
  year={2025}
}

@inproceedings{skand2025masked,
  title={Simple Masked Training Strategies Yield Control Policies That Are Robust to Sensor Failure},
  author={Skand, Skand and Pandit, Bikram and Kim, Chanho and Fuxin, Li and Lee, Stefan},
  booktitle={Proceedings of the 8th Conference on Robot Learning (CoRL)},
  series={Proceedings of Machine Learning Research},
  volume={270},
  pages={4463--4482},
  year={2025},
  publisher={PMLR}
}

@article{vogtlowell2026sensors,
  title={When Sensors Fail: Temporal Sequence Models for Robust PPO under Sensor Drift},
  author={Vogt-Lowell, Kevin and Tsiligkaridis, Theodoros and Lafuente-Mercado, Rodney and Ghatti, Surabhi and Gao, Shanghua and Zitnik, Marinka and Rus, Daniela},
  journal={arXiv preprint arXiv:2603.04648},
  year={2026},
  note={ICLR 2026 CAO Workshop}
}

@article{disdp2025,
  title={{DisDP}: Robust Imitation Learning via Disentangled Diffusion Policies},
  author={Vanjani, Pankhuri and Mattes, Paul and Jia, Xiaogang and Dave, Vedant and Lioutikov, Rudolf},
  journal={Reinforcement Learning Journal},
  volume={6},
  pages={1180--1199},
  year={2025}
}

@article{rl4il2026,
  title={Reinforcement Learning-Guided Retrieval with Soft Fusion for Robust Multimodal Imitation Learning under Missing Modalities},
  author={Ismkhan, Hassan and Bouchahcia, Hamid},
  journal={arXiv preprint arXiv:2606.15514},
  year={2026}
}

@inproceedings{robopanoptes2025,
  title={{RoboPanoptes}: The All-Seeing Robot with Whole-body Dexterity},
  author={Xu, Xiaomeng and Bauer, Dominik and Song, Shuran},
  booktitle={Robotics: Science and Systems (RSS)},
  year={2025},
  doi={10.15607/RSS.2025.XXI.042}
}

@inproceedings{wang2023shaspec,
  title={Multi-Modal Learning with Missing Modality via Shared-Specific Feature Modelling},
  author={Wang, Hu and Chen, Yuanhong and Ma, Congbo and Avery, Jodie and Hull, Louise and Carneiro, Gustavo},
  booktitle={Proceedings of the IEEE/CVF Conference on Computer Vision and Pattern Recognition (CVPR)},
  pages={15878--15887},
  year={2023},
  doi={10.1109/CVPR52729.2023.01524}
}

@inproceedings{kim2026must,
  title={{MUST}: Modality-Specific Representation-Aware Transformer for Diffusion-Enhanced Survival Prediction with Missing Modality},
  author={Kim, Kyungwon and Hwang, Dosik},
  booktitle={Proceedings of the IEEE/CVF Conference on Computer Vision and Pattern Recognition (CVPR)},
  pages={30312--30321},
  year={2026}
}

@inproceedings{xu2025dispro,
  title={Distilled Prompt Learning for Incomplete Multimodal Survival Prediction},
  author={Xu, Yingxue and Zhou, Fengtao and Zhao, Chenyu and Wang, Yihui and Yang, Can and Chen, Hao},
  booktitle={Proceedings of the IEEE/CVF Conference on Computer Vision and Pattern Recognition (CVPR)},
  pages={5102--5111},
  year={2025},
  doi={10.1109/CVPR52734.2025.00481}
}

@inproceedings{guo2025ctrlworld,
  title={{Ctrl-World}: A Controllable Generative World Model for Robot Manipulation},
  author={Guo, Yanjiang and Shi, Lucy Xiaoyang and Chen, Jianyu and Finn, Chelsea},
  booktitle={International Conference on Learning Representations (ICLR)},
  year={2026},
  note={arXiv:2510.10125}
}

@inproceedings{persistworld2026,
  title={Persistent Robot World Models: Stabilizing Multi-Step Rollouts via Reinforcement Learning},
  author={Bardhan, Jai and Drozdik, Patrik and Sivic, Josef and Petrik, Vladimir},
  booktitle={European Conference on Computer Vision (ECCV)},
  year={2026},
  note={arXiv:2603.25685}
}

@inproceedings{farneback2003,
  title={Two-Frame Motion Estimation Based on Polynomial Expansion},
  author={Farneb{\"a}ck, Gunnar},
  booktitle={Image Analysis (SCIA)},
  series={Lecture Notes in Computer Science},
  volume={2749},
  pages={363--370},
  year={2003},
  publisher={Springer},
  doi={10.1007/3-540-45103-X_50}
}

@misc{openpi2024,
  title={openpi: Open-Source Models and Packages for Robotics},
  author={{Physical Intelligence}},
  year={2025},
  howpublished={\url{https://github.com/Physical-Intelligence/openpi}}
}

@misc{nvidia2025isaacgroot,
  title={{NVIDIA} {Isaac} {GR00T}: A Foundation Model for Generalist Robots},
  author={{NVIDIA}},
  year={2025},
  howpublished={\url{https://github.com/NVIDIA/Isaac-GR00T}}
}

@article{brohan2023rt2,
  title   = {{RT-2}: Vision-Language-Action Models Transfer Web Knowledge to Robotic Control},
  author  = {Brohan, Anthony and Brown, Noah and Carbajal, Justice and Chebotar, Yevgen and Chen, Xi and Choromanski, Krzysztof and Ding, Tianli and Driess, Danny and Dubey, Avinava and Finn, Chelsea and others},
  journal = {arXiv preprint arXiv:2307.15818},
  year    = {2023}
}

@inproceedings{kim2024openvla,
  title     = {{OpenVLA}: An Open-Source Vision-Language-Action Model},
  author    = {Kim, Moo Jin and Pertsch, Karl and Karamcheti, Siddharth and Xiao, Ted and Balakrishna, Ashwin and Nair, Suraj and Rafailov, Rafael and Foster, Ethan P. and Sanketi, Pannag R. and Vuong, Quan and others},
  booktitle = {Proceedings of the 8th Conference on Robot Learning (CoRL)},
  series    = {Proceedings of Machine Learning Research},
  volume    = {270},
  pages     = {2679--2713},
  year      = {2025},
  publisher = {PMLR}
}

@inproceedings{octo2024,
  title     = {Octo: An Open-Source Generalist Robot Policy},
  author    = {{Octo Model Team} and Ghosh, Dibya and Walke, Homer and Pertsch, Karl and Black, Kevin and Mees, Oier and Dasari, Sudeep and Hejna, Joey and Kreiman, Tobias and Xu, Charles and others},
  booktitle = {Robotics: Science and Systems (RSS)},
  year      = {2024},
  doi       = {10.15607/RSS.2024.XX.090}
}

@article{xu2026stellavla,
  title   = {{StellaVLA}: In-Context Structured Demonstration for Generalizable Vision-Language-Action Models},
  author  = {Xu, Siyu and Wang, Yunke and Wang, Zijian and Zhu, Dihao and Xia, Chenghao and Du, Chengbin and Liu, Daochang and Huang, Tao and Xu, Chang},
  journal = {arXiv preprint arXiv:2608.11671},
  year    = {2026}
}
\bibliographystyle{iclr2027_conference}
\par}

\newpage
\appendix
\raggedbottom
\section{\mail{} details}
\label{app:mail-details}

\paragraph{Suite and horizons.}
The suite fixes the 18 Atomic-Seen tasks, 50 scenes per task and their horizons
(Table~\ref{tab:suite}) under RoboCasa365~1.0.1.

\begin{table}[H]
\centering\small
\caption{\textbf{Task suite.} The 18 RoboCasa365 Atomic-Seen tasks and their
horizons in control steps (20\,Hz).}
\label{tab:suite}
\begin{tabular}{@{}lr@{\hspace{2em}}lr@{}}
\toprule
Task & Horizon & Task & Horizon \\
\midrule
CloseBlenderLid & 900 & PickPlaceCounterToStove & 600 \\
CloseFridge & 900 & PickPlaceDrawerToCounter & 750 \\
CloseToasterOvenDoor & 450 & PickPlaceSinkToCounter & 900 \\
CoffeeSetupMug & 600 & PickPlaceToasterToCounter & 600 \\
NavigateKitchen & 450 & SlideDishwasherRack & 450 \\
OpenCabinet & 1050 & TurnOffStove & 750 \\
OpenDrawer & 750 & TurnOnElectricKettle & 450 \\
OpenStandMixerHead & 450 & TurnOnMicrowave & 450 \\
PickPlaceCounterToCabinet & 750 & TurnOnSinkFaucet & 600 \\
\bottomrule
\end{tabular}
\end{table}

\paragraph{Base policies.}
We evaluate the RoboCasa365 Human300-trained checkpoints of $\pi_{0.5}$
\citep{black2025pi05} and GR00T~N1.5 \citep{nvidia2025groot}, served through
their native runtimes, openpi \citep{openpi2024} and Isaac-GR00T
\citep{nvidia2025isaacgroot}. Under healthy observations, $\pi_{0.5}$ solves 362 of the 900
scenes ($H=0.402$; 0.396 on the RoboCasa365 leaderboard) and GR00T~N1.5 solves
405 ($H=0.450$; 0.430 in the original paper and 0.507 on the leaderboard,
re-evaluated under the 1.5$\times$ RoboCasa365~1.0.1 horizons that \mail{} pins).
Missing-view training does not reduce healthy success: on GR00T~N1.5,
\emph{Clean FT + prediction}, \emph{Missing-view FT} and \mint{} reach
$H=0.532$, $0.540$ and $0.544$ (Table~\ref{tab:mail-ladder}).

\paragraph{Auxiliary fault operators.}
The released kernel also implements blackout, freeze, stale delivery, burst
dropout, finite-duration loss and recovery for diagnostic use; these are not
part of the MAIL-Bench official score.

\paragraph{Determinism.}
Policy sampling is seeded from the benchmark seed and the simulator is reset to
the frozen scene state before every rollout, so paired rollouts differ only
through the camera intervention, up to small GPU-renderer differences that
\mail{} records.

\section{\mint{} Implementation Details}
\label{app:mint-details}

\subsection{Missing-view training}
\label{app:groot}

We describe the GR00T~N1.5 implementation; $\pi_{0.5}$ follows the same recipe
within its native backbone. Training uses the curriculum $\mathcal{C}$ of
Eq.~\ref{eq:train} with the values in Table~\ref{tab:mint-constants}: a faulted
episode loses a camera group either persistently from a random onset or in
short intermittent bursts, and the two third-person cameras always share one
availability state.

Unavailable cameras are represented exactly as in the main method: their
availability flag is 0, their image slot contains an inert zero placeholder,
and their visual tokens are excluded from policy computation through the
attention mask
\begin{equation}
\label{eq:mask}
\mathrm{Attn}(Q,K,V)=
\mathrm{softmax}\!\Big(
\tfrac{QK^{\top}}{\sqrt{d}}+\log M
\Big)V,
\quad
M_{ij}=
\begin{cases}
m^{\,r(j)} & j\ \text{a visual token of group } r(j),\\
1 & \text{otherwise},
\end{cases}
\end{equation}
with $\log 0=-\infty$, so an unavailable camera contributes zero attention
weight while an available camera is processed as usual; changing the pixels of
a masked camera leaves the actions unchanged.

\subsection{Serving and prediction routes}
\label{app:serving}

\paragraph{Withdrawal.}
Withdrawal follows Eq.~\ref{eq:gate}: once a route fails a check, the role
returns to the missing input used in training for the rest of the episode. The
optical-flow route is checked only after a minimum tenancy and has no upper
limit on its duration; its propagated-view consistency is a flow confidence,
the product of a photometric-consistency term, an accumulated-drift term and
the fraction of pixels with a valid warp source. The world-model route checks
every generated block from the first and is withdrawn at the first rejected
block or at the generation cap. The two third-person cameras are admitted and
withdrawn together.

\paragraph{World-model route.}
We use PersistWorld because it is action-conditioned and sustains multi-step
rollouts within the serving budget, and adapt it to the 18 tasks with a
low-rank fine-tune on the demonstrations. On held-out demonstrations,
adaptation raises reconstruction PSNR from 16.4 to 18.9\,dB and lowers the
fraction of out-of-domain conditioning states from 0.18 to 0.01. The rollout
memory $\mathcal{M}$ of Eq.~\ref{eq:wm} is initialized from the last available
observations, and the auxiliary wrist prediction is used only for the
consistency check. The same adaptation is used on the AgiBot~G2
(Appendix~\ref{app:g2}).

\subsection{Choice of Hyperparameters}
\label{app:constants}

All hyperparameters were fixed on development scenes disjoint from the 900
benchmark scenes, before benchmark evaluation (Table~\ref{tab:mint-constants}).

\begin{table}[H]
\centering
\small
\caption{\textbf{Hyperparameters of \mint{}.}}
\label{tab:mint-constants}
\begin{tabular}{@{}ll@{}}
\toprule
Hyperparameter & Value \\
\midrule
\multicolumn{2}{@{}l}{\textit{Optical-flow route}} \\
Min.\ flow confidence (wrist / third-person) & 0.05 / 0.08 \\
Motion budget, wrist (end-effector + base) & 0.15\,m / 0.70\,rad \\
Motion budget, third-person (base) & 0.30\,m / 0.70\,rad \\
Minimum tenancy & 64 steps \\
\midrule
\multicolumn{2}{@{}l}{\textit{World-model route}} \\
Max.\ wrist consistency error $d_q^w$ & 0.20 \\
Visual change (max / min while moving) & 0.35 / 0.002 \\
Max.\ out-of-domain conditioning fraction & 0.35 \\
Generation cap & 8 blocks (64 steps) \\
\midrule
\multicolumn{2}{@{}l}{\textit{Training and serving}} \\
Curriculum $\mathcal{C}$ (healthy / wrist / third-person / all vision) & 0.60 / 0.17 / 0.17 / 0.06 \\
Persistent-interruption probability & 0.7 \\
Persistent onset (fraction of episode) & $\mathcal{U}[0.3,0.7]$ \\
Fine-tuning steps & 30k \\
Query interval & 8 steps \\
Action horizon $h$ (GR00T~N1.5 / $\pi_{0.5}$) & 16 / 50 \\
\bottomrule
\end{tabular}
\end{table}

\FloatBarrier
\section{\mail{} per-task and per-condition results}
\label{app:mail-tables}
All tables are generated from the scorer output of runs with complete healthy
and fault coverage and no audit failure; Table~\ref{tab:mail-ci} gives paired
bootstrap intervals. The task categories of
Table~\ref{tab:mail-grid-categories} are pick-and-place (the five PickPlace
tasks), small appliances (CloseBlenderLid, CoffeeSetupMug, OpenStandMixerHead,
SlideDishwasherRack, TurnOnElectricKettle, TurnOnMicrowave) and fixtures and
navigation (CloseFridge, CloseToasterOvenDoor, NavigateKitchen, OpenCabinet,
OpenDrawer, TurnOffStove, TurnOnSinkFaucet). In the per-task tables, $N^H_j$ is
the number of healthy successes out of 50 scenes and ``n/a'' marks a task
without healthy success, which contributes zero to $S_j$; values with small
$N^H_j$ rest on few scenes.

\begin{table}[!htb]
\caption{\textbf{Main results.} Healthy success $H$, conditional scores $M_c$
under wrist (W), third-person (T) and all-vision (B) interruption at the
30/45/60\% onsets, and $S_{\mathrm{MAIL}}$ (Eq.~\ref{eq:score}), averaged over
the 18 tasks.}
\label{tab:mail}
\centering\footnotesize
\setlength{\tabcolsep}{3.2pt}
\resizebox{\textwidth}{!}{
\begin{tabular}{lc ccc ccc ccc c}
\toprule
& & \multicolumn{3}{c}{Wrist missing} & \multicolumn{3}{c}{Third-person missing} & \multicolumn{3}{c}{All vision missing} & \\
\cmidrule(lr){3-5}\cmidrule(lr){6-8}\cmidrule(lr){9-11}
Policy & $H$ & W30 & W45 & W60 & T30 & T45 & T60 & B30 & B45 & B60 & $S_{\mathrm{MAIL}}$ \\
\midrule
$\pi_{0.5}$ & 0.402 & 0.165 & 0.219 & 0.307 & 0.340 & 0.365 & 0.476 & 0.038 & 0.086 & 0.176 & \textbf{0.257} \\
\rowcolor{black!8}$\pi_{0.5}$ + \mint{} & 0.463 & 0.156 & 0.274 & 0.417 & 0.497 & 0.522 & 0.571 & 0.116 & 0.187 & 0.286 & \textbf{0.349} \\
GR00T N1.5 & 0.450 & 0.165 & 0.267 & 0.377 & 0.247 & 0.285 & 0.407 & 0.048 & 0.089 & 0.124 & \textbf{0.246} \\
\rowcolor{black!8}GR00T N1.5 + \mint{} & 0.544 & 0.457 & 0.532 & 0.650 & 0.504 & 0.586 & 0.591 & 0.147 & 0.242 & 0.369 & \textbf{0.462} \\
\bottomrule
\end{tabular}}
\end{table}

\begin{table}[t]
\caption{\textbf{GR00T~N1.5 ablation.} All variants except the base are served
at the same eight-step cadence. \emph{Clean FT + prediction} is fine-tuned on
the same data and steps as \mint{} without missing-view examples;
\emph{Missing-view FT} is the \mint{} checkpoint without generated views.
$\Delta$ is $S_{\mathrm{MAIL}}$ minus the row above.}
\label{tab:mail-ladder}
\centering\footnotesize
\setlength{\tabcolsep}{3.0pt}
\resizebox{\textwidth}{!}{
\begin{tabular}{lc ccc ccc ccc cc}
\toprule
& & \multicolumn{3}{c}{Wrist missing} & \multicolumn{3}{c}{Third-person missing} & \multicolumn{3}{c}{All vision missing} & & \\
\cmidrule(lr){3-5}\cmidrule(lr){6-8}\cmidrule(lr){9-11}
Arm & $H$ & W30 & W45 & W60 & T30 & T45 & T60 & B30 & B45 & B60 & $S_{\mathrm{MAIL}}$ & $\Delta$ \\
\midrule
Base & 0.450 & 0.165 & 0.267 & 0.377 & 0.247 & 0.285 & 0.407 & 0.048 & 0.089 & 0.124 & \textbf{0.246} &  \\
Clean FT + prediction & 0.532 & 0.294 & 0.369 & 0.580 & 0.316 & 0.419 & 0.455 & 0.079 & 0.191 & 0.279 & \textbf{0.351} & +0.105 \\
Missing-view FT & 0.540 & 0.412 & 0.542 & 0.643 & 0.405 & 0.495 & 0.605 & 0.151 & 0.252 & 0.334 & \textbf{0.438} & +0.086 \\
\rowcolor{black!8}\mint{} & 0.544 & 0.457 & 0.532 & 0.650 & 0.504 & 0.586 & 0.591 & 0.147 & 0.242 & 0.369 & \textbf{0.462} & +0.024 \\
\bottomrule
\end{tabular}}
\end{table}

\begin{table}[!htb]
\caption{\textbf{Bootstrap 95\% intervals.} Scenes are resampled within each
task (10{,}000 resamples), and differences between arms are paired on the same
resampled scenes. Top: $S_{\mathrm{MAIL}}$ and paired differences. Bottom:
per-condition paired differences between each backbone with \mint{} and its
base.}
\label{tab:mail-ci}
\centering\footnotesize
\setlength{\tabcolsep}{4pt}
\resizebox{\textwidth}{!}{
\begin{tabular}{@{}l c c@{\hspace{2.2em}}l c c@{}}
\toprule
Arm & $S_{\mathrm{MAIL}}$ & 95\% interval & Paired difference & $\Delta S_{\mathrm{MAIL}}$ & 95\% interval \\
\midrule
$\pi_{0.5}$ & 0.257 & [0.221, 0.275] & $\pi_{0.5}$ + \mint{} $-$ $\pi_{0.5}$ & $+0.091$ & $[+0.063, +0.129]$ \\
$\pi_{0.5}$ + \mint{} & 0.349 & [0.325, 0.365] & GR00T N1.5 + \mint{} $-$ GR00T N1.5 & $+0.216$ & $[+0.190, +0.240]$ \\
GR00T N1.5 & 0.246 & [0.227, 0.265] & Clean FT + prediction $-$ GR00T N1.5 & $+0.105$ & $[+0.079, +0.130]$ \\
Clean FT + prediction & 0.351 & [0.332, 0.369] & Missing-view FT $-$ Clean FT + prediction & $+0.086$ & $[+0.061, +0.109]$ \\
Missing-view FT & 0.438 & [0.417, 0.455] & GR00T N1.5 + \mint{} $-$ Missing-view FT & $+0.024$ & $[+0.008, +0.044]$ \\
GR00T N1.5 + \mint{} & 0.462 & [0.443, 0.479] & GR00T N1.5 + \mint{} $-$ Clean FT + prediction & $+0.111$ & $[+0.087, +0.133]$ \\
\bottomrule
\end{tabular}}
\\[8pt]
\begin{tabular}{@{}l c c c c@{}}
\toprule
 & \multicolumn{2}{c}{$\pi_{0.5}$ + \mint{} $-$ $\pi_{0.5}$} & \multicolumn{2}{c}{GR00T N1.5 + \mint{} $-$ GR00T N1.5} \\
\cmidrule(lr){2-3}\cmidrule(lr){4-5}
Condition & $\Delta M_c$ & 95\% interval & $\Delta M_c$ & 95\% interval \\
\midrule
W30 & $-0.010$ & $[-0.068, +0.044]$ & $+0.291$ & $[+0.235, +0.345]$ \\
W45 & $+0.056$ & $[-0.022, +0.145]$ & $+0.264$ & $[+0.206, +0.321]$ \\
W60 & $+0.110$ & $[+0.018, +0.180]$ & $+0.273$ & $[+0.208, +0.344]$ \\
T30 & $+0.157$ & $[+0.079, +0.237]$ & $+0.257$ & $[+0.179, +0.331]$ \\
T45 & $+0.158$ & $[+0.095, +0.226]$ & $+0.301$ & $[+0.223, +0.372]$ \\
T60 & $+0.095$ & $[+0.043, +0.175]$ & $+0.184$ & $[+0.114, +0.249]$ \\
B30 & $+0.078$ & $[+0.042, +0.113]$ & $+0.099$ & $[+0.067, +0.131]$ \\
B45 & $+0.101$ & $[+0.046, +0.159]$ & $+0.152$ & $[+0.105, +0.196]$ \\
B60 & $+0.110$ & $[+0.065, +0.185]$ & $+0.245$ & $[+0.198, +0.292]$ \\
\bottomrule
\end{tabular}
\end{table}

\begin{table}[!ht]
\caption{Per-task, per-condition results for $\pi_{0.5}$.}
\label{tab:mail-cond-pi05}
\centering\footnotesize
\setlength{\tabcolsep}{3pt}
\resizebox{\textwidth}{!}{
\begin{tabular}{lc ccc ccc ccc c}
\toprule
Task & $N^H_j$ & W30 & W45 & W60 & T30 & T45 & T60 & B30 & B45 & B60 & $S_j$ \\
\midrule
CloseBlenderLid & 0 & n/a & n/a & n/a & n/a & n/a & n/a & n/a & n/a & n/a & 0.000 \\
CloseFridge & 23 & 0.70 & 0.57 & 0.74 & 0.09 & 0.17 & 0.39 & 0.09 & 0.26 & 0.57 & 0.403 \\
CloseToasterOvenDoor & 2 & 0.00 & 1.00 & 1.00 & 0.50 & 0.50 & 1.00 & 0.00 & 0.50 & 1.00 & 0.554 \\
CoffeeSetupMug & 5 & 0.00 & 0.00 & 0.00 & 0.20 & 0.20 & 0.20 & 0.00 & 0.00 & 0.00 & 0.070 \\
NavigateKitchen & 0 & n/a & n/a & n/a & n/a & n/a & n/a & n/a & n/a & n/a & 0.000 \\
OpenCabinet & 34 & 0.03 & 0.12 & 0.18 & 0.09 & 0.15 & 0.15 & 0.00 & 0.00 & 0.00 & 0.139 \\
OpenDrawer & 28 & 0.00 & 0.04 & 0.18 & 0.61 & 0.75 & 0.82 & 0.00 & 0.00 & 0.00 & 0.295 \\
OpenStandMixerHead & 27 & 0.81 & 0.85 & 0.96 & 0.56 & 0.63 & 0.67 & 0.11 & 0.19 & 0.22 & 0.554 \\
PickPlaceCounterToCabinet & 39 & 0.05 & 0.13 & 0.15 & 0.00 & 0.00 & 0.13 & 0.00 & 0.00 & 0.03 & 0.127 \\
PickPlaceCounterToStove & 38 & 0.16 & 0.18 & 0.26 & 0.45 & 0.55 & 0.87 & 0.00 & 0.03 & 0.26 & 0.352 \\
PickPlaceDrawerToCounter & 23 & 0.04 & 0.04 & 0.13 & 0.09 & 0.09 & 0.17 & 0.00 & 0.04 & 0.09 & 0.116 \\
PickPlaceSinkToCounter & 45 & 0.04 & 0.13 & 0.31 & 0.44 & 0.44 & 0.49 & 0.00 & 0.00 & 0.00 & 0.277 \\
PickPlaceToasterToCounter & 14 & 0.00 & 0.00 & 0.14 & 0.21 & 0.14 & 0.21 & 0.00 & 0.00 & 0.07 & 0.107 \\
SlideDishwasherRack & 32 & 0.50 & 0.50 & 0.62 & 0.53 & 0.69 & 0.84 & 0.12 & 0.16 & 0.56 & 0.517 \\
TurnOffStove & 9 & 0.33 & 0.11 & 0.22 & 0.89 & 0.78 & 0.89 & 0.11 & 0.11 & 0.11 & 0.374 \\
TurnOnElectricKettle & 13 & 0.31 & 0.23 & 0.46 & 0.46 & 0.54 & 0.77 & 0.15 & 0.23 & 0.15 & 0.357 \\
TurnOnMicrowave & 0 & n/a & n/a & n/a & n/a & n/a & n/a & n/a & n/a & n/a & 0.000 \\
TurnOnSinkFaucet & 30 & 0.00 & 0.03 & 0.17 & 1.00 & 0.93 & 0.97 & 0.10 & 0.03 & 0.10 & 0.393 \\
\midrule
Mean over 18 tasks & 362 & 0.165 & 0.219 & 0.307 & 0.340 & 0.365 & 0.476 & 0.038 & 0.086 & 0.176 & \textbf{0.257} \\
\bottomrule
\end{tabular}}
\end{table}

\begin{table}[!ht]
\caption{Per-task, per-condition results for $\pi_{0.5}$ + \mint{}.}
\label{tab:mail-cond-pi05-p1control-mint}
\centering\footnotesize
\setlength{\tabcolsep}{3pt}
\resizebox{\textwidth}{!}{
\begin{tabular}{lc ccc ccc ccc c}
\toprule
Task & $N^H_j$ & W30 & W45 & W60 & T30 & T45 & T60 & B30 & B45 & B60 & $S_j$ \\
\midrule
CloseBlenderLid & 2 & 0.00 & 0.50 & 0.00 & 0.50 & 0.00 & 0.00 & 0.00 & 0.00 & 0.00 & 0.104 \\
CloseFridge & 23 & 0.70 & 0.74 & 0.78 & 0.48 & 0.48 & 0.70 & 0.26 & 0.22 & 0.70 & 0.550 \\
CloseToasterOvenDoor & 9 & 0.11 & 0.22 & 0.56 & 0.11 & 0.22 & 0.33 & 0.22 & 0.22 & 0.33 & 0.251 \\
CoffeeSetupMug & 13 & 0.00 & 0.00 & 0.00 & 0.46 & 0.38 & 0.69 & 0.00 & 0.00 & 0.15 & 0.195 \\
NavigateKitchen & 1 & 0.00 & 0.00 & 1.00 & 0.00 & 0.00 & 0.00 & 0.00 & 0.00 & 0.00 & 0.102 \\
OpenCabinet & 36 & 0.06 & 0.22 & 0.31 & 0.11 & 0.36 & 0.25 & 0.00 & 0.03 & 0.03 & 0.208 \\
OpenDrawer & 31 & 0.00 & 0.16 & 0.45 & 0.58 & 0.74 & 0.81 & 0.03 & 0.13 & 0.26 & 0.378 \\
OpenStandMixerHead & 37 & 0.38 & 0.62 & 0.86 & 0.70 & 0.65 & 0.78 & 0.59 & 0.78 & 0.86 & 0.698 \\
PickPlaceCounterToCabinet & 39 & 0.05 & 0.15 & 0.62 & 0.67 & 0.87 & 0.79 & 0.03 & 0.21 & 0.46 & 0.463 \\
PickPlaceCounterToStove & 36 & 0.00 & 0.17 & 0.44 & 0.69 & 0.94 & 0.67 & 0.00 & 0.06 & 0.14 & 0.383 \\
PickPlaceDrawerToCounter & 26 & 0.04 & 0.08 & 0.23 & 0.27 & 0.38 & 0.35 & 0.08 & 0.12 & 0.31 & 0.237 \\
PickPlaceSinkToCounter & 44 & 0.05 & 0.05 & 0.27 & 0.50 & 0.77 & 0.80 & 0.02 & 0.00 & 0.00 & 0.333 \\
PickPlaceToasterToCounter & 27 & 0.00 & 0.11 & 0.15 & 0.74 & 0.70 & 0.96 & 0.00 & 0.07 & 0.07 & 0.335 \\
SlideDishwasherRack & 32 & 0.66 & 0.72 & 0.88 & 0.84 & 0.84 & 0.91 & 0.50 & 0.81 & 0.84 & 0.764 \\
TurnOffStove & 8 & 0.00 & 0.38 & 0.25 & 0.88 & 0.88 & 0.88 & 0.00 & 0.00 & 0.25 & 0.366 \\
TurnOnElectricKettle & 14 & 0.21 & 0.21 & 0.36 & 0.57 & 0.36 & 0.50 & 0.21 & 0.43 & 0.36 & 0.349 \\
TurnOnMicrowave & 2 & 0.50 & 0.50 & 0.00 & 0.00 & 0.00 & 0.00 & 0.00 & 0.00 & 0.00 & 0.104 \\
TurnOnSinkFaucet & 37 & 0.05 & 0.11 & 0.35 & 0.84 & 0.81 & 0.86 & 0.14 & 0.30 & 0.38 & 0.458 \\
\midrule
Mean over 18 tasks & 417 & 0.156 & 0.274 & 0.417 & 0.497 & 0.522 & 0.571 & 0.116 & 0.187 & 0.286 & \textbf{0.349} \\
\bottomrule
\end{tabular}}
\end{table}

\begin{table}[!ht]
\caption{Per-task, per-condition results for GR00T N1.5.}
\label{tab:mail-cond-groot15}
\centering\footnotesize
\setlength{\tabcolsep}{3pt}
\resizebox{\textwidth}{!}{
\begin{tabular}{lc ccc ccc ccc c}
\toprule
Task & $N^H_j$ & W30 & W45 & W60 & T30 & T45 & T60 & B30 & B45 & B60 & $S_j$ \\
\midrule
CloseBlenderLid & 3 & 0.00 & 0.00 & 0.67 & 0.00 & 0.00 & 0.00 & 0.00 & 0.00 & 0.00 & 0.073 \\
CloseFridge & 35 & 0.57 & 0.71 & 0.74 & 0.17 & 0.31 & 0.37 & 0.20 & 0.29 & 0.46 & 0.453 \\
CloseToasterOvenDoor & 12 & 0.25 & 0.50 & 0.92 & 0.08 & 0.17 & 0.42 & 0.00 & 0.00 & 0.08 & 0.266 \\
CoffeeSetupMug & 5 & 0.00 & 0.00 & 0.00 & 0.00 & 0.00 & 0.20 & 0.00 & 0.00 & 0.00 & 0.030 \\
NavigateKitchen & 5 & 0.20 & 0.20 & 0.20 & 0.40 & 0.00 & 0.00 & 0.00 & 0.20 & 0.00 & 0.130 \\
OpenCabinet & 32 & 0.19 & 0.25 & 0.44 & 0.06 & 0.06 & 0.16 & 0.00 & 0.03 & 0.00 & 0.183 \\
OpenDrawer & 28 & 0.04 & 0.11 & 0.29 & 0.54 & 0.54 & 0.61 & 0.00 & 0.07 & 0.18 & 0.292 \\
OpenStandMixerHead & 26 & 0.31 & 0.58 & 0.77 & 0.27 & 0.42 & 0.69 & 0.19 & 0.23 & 0.31 & 0.429 \\
PickPlaceCounterToCabinet & 41 & 0.46 & 0.61 & 0.54 & 0.17 & 0.22 & 0.49 & 0.00 & 0.00 & 0.00 & 0.331 \\
PickPlaceCounterToStove & 41 & 0.02 & 0.20 & 0.41 & 0.12 & 0.22 & 0.46 & 0.00 & 0.00 & 0.02 & 0.228 \\
PickPlaceDrawerToCounter & 21 & 0.00 & 0.05 & 0.00 & 0.19 & 0.24 & 0.48 & 0.00 & 0.00 & 0.00 & 0.137 \\
PickPlaceSinkToCounter & 41 & 0.07 & 0.24 & 0.32 & 0.32 & 0.32 & 0.51 & 0.00 & 0.00 & 0.02 & 0.262 \\
PickPlaceToasterToCounter & 39 & 0.26 & 0.28 & 0.44 & 0.03 & 0.13 & 0.18 & 0.00 & 0.00 & 0.00 & 0.209 \\
SlideDishwasherRack & 24 & 0.38 & 0.62 & 0.58 & 0.79 & 0.71 & 0.79 & 0.33 & 0.38 & 0.54 & 0.560 \\
TurnOffStove & 5 & 0.00 & 0.00 & 0.00 & 0.40 & 0.60 & 0.40 & 0.00 & 0.00 & 0.00 & 0.150 \\
TurnOnElectricKettle & 22 & 0.09 & 0.32 & 0.23 & 0.41 & 0.50 & 0.59 & 0.14 & 0.27 & 0.36 & 0.335 \\
TurnOnMicrowave & 7 & 0.14 & 0.14 & 0.14 & 0.00 & 0.14 & 0.14 & 0.00 & 0.14 & 0.14 & 0.114 \\
TurnOnSinkFaucet & 18 & 0.00 & 0.00 & 0.11 & 0.50 & 0.56 & 0.83 & 0.00 & 0.00 & 0.11 & 0.247 \\
\midrule
Mean over 18 tasks & 405 & 0.165 & 0.267 & 0.377 & 0.247 & 0.285 & 0.407 & 0.048 & 0.089 & 0.124 & \textbf{0.246} \\
\bottomrule
\end{tabular}}
\end{table}

\begin{table}[!ht]
\caption{Per-task, per-condition results for GR00T N1.5 + \mint{}.}
\label{tab:mail-cond-groot15-p1control-mint}
\centering\footnotesize
\setlength{\tabcolsep}{3pt}
\resizebox{\textwidth}{!}{
\begin{tabular}{lc ccc ccc ccc c}
\toprule
Task & $N^H_j$ & W30 & W45 & W60 & T30 & T45 & T60 & B30 & B45 & B60 & $S_j$ \\
\midrule
CloseBlenderLid & 2 & 0.00 & 0.00 & 0.00 & 0.50 & 0.50 & 0.00 & 0.00 & 0.00 & 0.00 & 0.104 \\
CloseFridge & 30 & 0.83 & 0.70 & 0.80 & 0.30 & 0.53 & 0.63 & 0.23 & 0.20 & 0.37 & 0.520 \\
CloseToasterOvenDoor & 19 & 0.26 & 0.37 & 0.63 & 0.37 & 0.26 & 0.37 & 0.00 & 0.05 & 0.32 & 0.301 \\
CoffeeSetupMug & 11 & 0.18 & 0.18 & 0.64 & 0.36 & 0.36 & 0.64 & 0.09 & 0.00 & 0.09 & 0.277 \\
NavigateKitchen & 14 & 0.43 & 0.57 & 0.50 & 0.00 & 0.07 & 0.14 & 0.00 & 0.00 & 0.07 & 0.207 \\
OpenCabinet & 41 & 0.78 & 0.73 & 0.78 & 0.22 & 0.39 & 0.54 & 0.05 & 0.07 & 0.15 & 0.453 \\
OpenDrawer & 36 & 0.28 & 0.58 & 0.81 & 0.64 & 0.75 & 0.81 & 0.08 & 0.31 & 0.78 & 0.575 \\
OpenStandMixerHead & 34 & 0.88 & 0.97 & 0.94 & 0.62 & 0.59 & 0.62 & 0.65 & 0.82 & 1.00 & 0.777 \\
PickPlaceCounterToCabinet & 42 & 0.71 & 0.79 & 0.88 & 0.83 & 0.90 & 0.76 & 0.24 & 0.38 & 0.57 & 0.691 \\
PickPlaceCounterToStove & 40 & 0.25 & 0.47 & 0.55 & 0.75 & 0.62 & 0.47 & 0.00 & 0.00 & 0.00 & 0.392 \\
PickPlaceDrawerToCounter & 19 & 0.05 & 0.05 & 0.21 & 0.21 & 0.32 & 0.53 & 0.00 & 0.00 & 0.11 & 0.185 \\
PickPlaceSinkToCounter & 42 & 0.33 & 0.57 & 0.88 & 0.74 & 0.79 & 0.79 & 0.00 & 0.02 & 0.02 & 0.498 \\
PickPlaceToasterToCounter & 38 & 0.76 & 0.92 & 0.84 & 0.58 & 0.74 & 0.68 & 0.03 & 0.05 & 0.13 & 0.550 \\
SlideDishwasherRack & 39 & 0.82 & 0.85 & 0.82 & 0.72 & 0.74 & 0.82 & 0.49 & 0.82 & 0.90 & 0.775 \\
TurnOffStove & 10 & 0.40 & 0.30 & 0.40 & 0.40 & 0.70 & 0.60 & 0.00 & 0.50 & 0.30 & 0.380 \\
TurnOnElectricKettle & 34 & 0.47 & 0.62 & 0.59 & 0.88 & 0.97 & 0.76 & 0.35 & 0.47 & 0.65 & 0.644 \\
TurnOnMicrowave & 15 & 0.60 & 0.60 & 0.80 & 0.53 & 0.60 & 0.60 & 0.27 & 0.40 & 0.53 & 0.523 \\
TurnOnSinkFaucet & 24 & 0.17 & 0.29 & 0.62 & 0.42 & 0.71 & 0.88 & 0.17 & 0.25 & 0.67 & 0.465 \\
\midrule
Mean over 18 tasks & 490 & 0.457 & 0.532 & 0.650 & 0.504 & 0.586 & 0.591 & 0.147 & 0.242 & 0.369 & \textbf{0.462} \\
\bottomrule
\end{tabular}}
\end{table}

\subsection{Compute cost}
\label{app:compute}
All experiments run on NVIDIA GeForce RTX 4090 GPUs (24\,GB). Evaluating one
arm on \mail{} takes 60--70 single-worker hours, and fine-tuning takes 12.9
GPU-hours for GR00T~N1.5 (30k steps, one GPU) and 15.1 GPU-hours for
$\pi_{0.5}$ (30k steps, two GPUs). A policy query costs about 5.0\,TFLOPs for
$\pi_{0.5}$ and 2.5\,TFLOPs for GR00T~N1.5. \mint{} does not increase the
per-step wall-clock time of healthy execution; under interruption, the
optical-flow route runs on the CPU, whereas the world model adds 9.2\,GiB of
GPU memory and about 100\,TFLOPs per generation, with 7.35 (GR00T~N1.5) and
7.72 ($\pi_{0.5}$) generations per third-person-missing cell on average.

\FloatBarrier
\section{Real-Robot Setup}
\label{app:g2}

\begin{figure}[H]
\centering
\begin{subfigure}{\textwidth}
\centering
\includegraphics[width=0.8\textwidth]{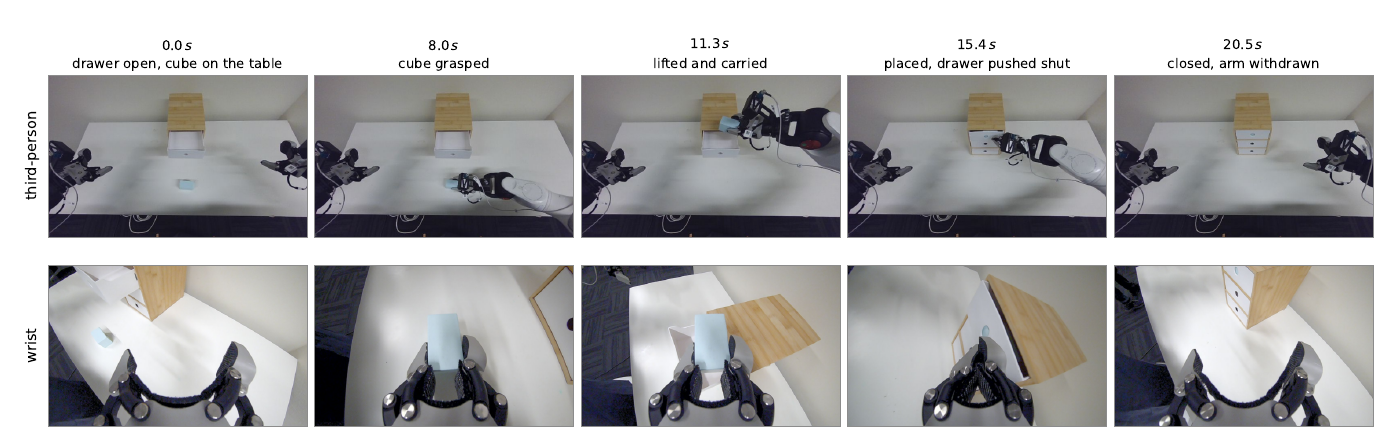}
\caption{Drawer.}
\label{fig:g2-drawer}
\end{subfigure}
\vspace{2pt}
\begin{subfigure}{\textwidth}
\centering
\includegraphics[width=0.8\textwidth]{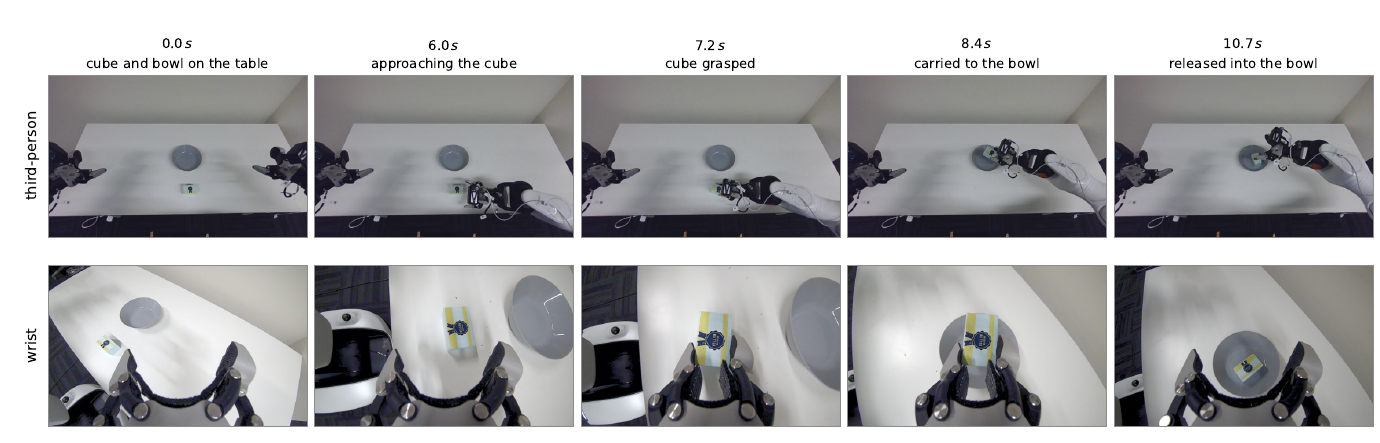}
\caption{Cube into bowl.}
\label{fig:g2-bowl}
\end{subfigure}
\vspace{2pt}
\begin{subfigure}{\textwidth}
\centering
\includegraphics[width=0.8\textwidth]{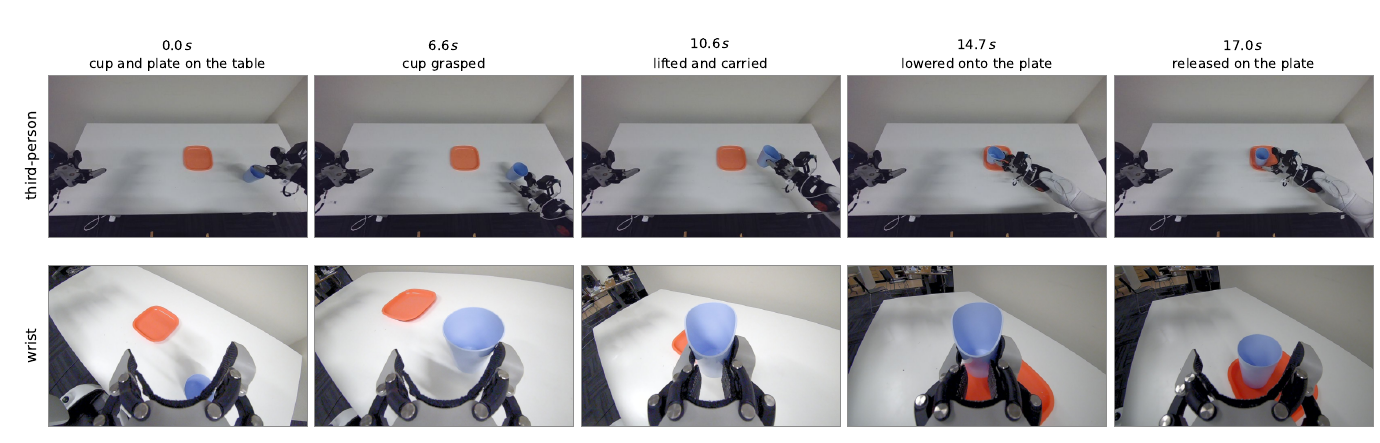}
\caption{Cup onto plate.}
\label{fig:g2-cup}
\end{subfigure}
\vspace{2pt}
\begin{subfigure}{\textwidth}
\centering
\includegraphics[width=0.8\textwidth]{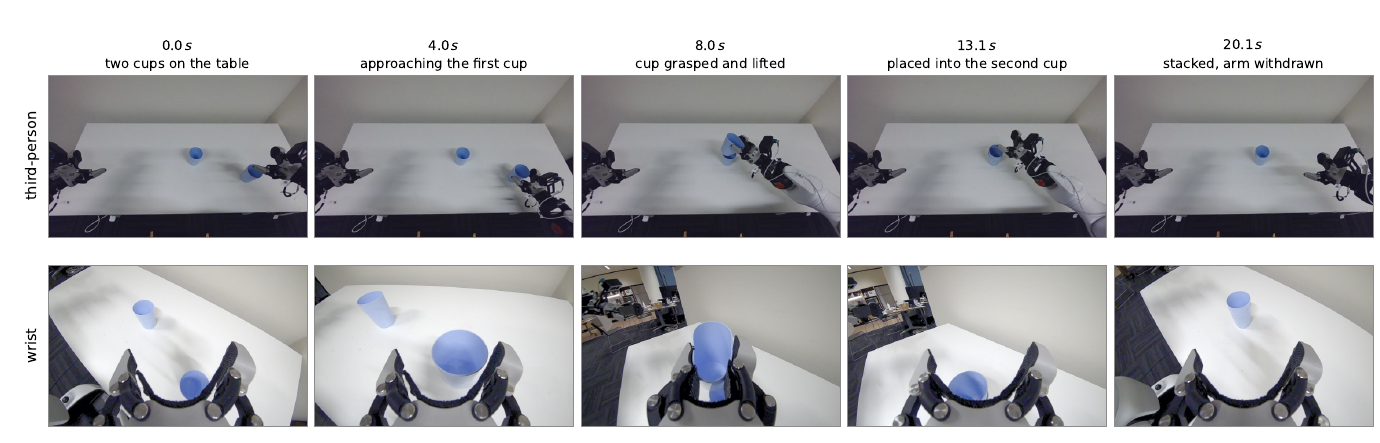}
\caption{Stack cups.}
\label{fig:g2-stack}
\end{subfigure}
\caption{\textbf{Healthy episodes of the four AgiBot~G2 tasks.} (a)~drawer,
(b)~cube into bowl, (c)~cup onto plate, (d)~stack cups. Each panel shows the head
(third-person) camera (top) and the right wrist camera (bottom) at five time
points.}
\label{fig:g2-episodes}
\end{figure}

The AgiBot~G2 has a head-mounted camera and one camera on each wrist, and the
right arm manipulates. The head camera supplies the third-person group and the
right wrist camera the wrist group. We collected 224 teleoperated episodes of
four tabletop tasks and fine-tune an availability-aware GR00T~N1.5 checkpoint
on the 203 successful ones. Figure~\ref{fig:g2-episodes} shows a healthy episode of each task.

\end{document}